\documentclass[lettersize,journal]{IEEEtran}
\usepackage{amsmath,amssymb,amsfonts}
\usepackage{algorithm}
\usepackage{algorithmic}
\usepackage{array}
\usepackage{subfigure}
\usepackage{textcomp}
\usepackage{stfloats}
\usepackage{url}
\usepackage{multirow}
\usepackage{verbatim}
\usepackage{booktabs}
\usepackage{graphicx}
\usepackage{hyperref}
\usepackage{threeparttable}
\newtheorem{definition}{Definition}

\def\BibTeX{{\rm B\kern-.05em{\sc i\kern-.025em b}\kern-.08em
    T\kern-.1667em\lower.7ex\hbox{E}\kern-.125emX}}
\usepackage{balance}
\begin{document}

\title{Teaching MLPs to Master Heterogeneous Graph-Structured Knowledge for Efficient and Accurate Inference}

\author{Yunhui~Liu, 
        Xinyi~Gao, 
        Tieke~He*,~\IEEEmembership{Member,~IEEE}, 
        Jianhua~Zhao, and\\
        Hongzhi~Yin,~\IEEEmembership{Senior Member,~IEEE}
        \IEEEcompsocitemizethanks{
        \IEEEcompsocthanksitem Yunhui~Liu, Tieke~He and Jianhua~Zhao are with the State Key Laboratory for Novel Software Technology, Nanjing University, Nanjing 210023, China (email: \{lyhcloudy1225, hetieke\}@gmail.com; zhaojh@nju.edu.cn). 
        \IEEEcompsocthanksitem Xinyi~Gao and Hongzhi~Yin are with the School of Information Technology and Electrical Engineering, The University of Queensland, Brisbane, Australia (email: \{xinyi.gao, h.yin1\}@uq.edu.au).}
        \thanks{*Corresponding author.}}


\markboth{Journal of \LaTeX\ Class Files,~Vol.~18, No.~9, September~2020}%
{How to Use the IEEEtran \LaTeX \ Templates}

\maketitle

\begin{abstract}
Heterogeneous Graph Neural Networks (HGNNs) have achieved promising results in various heterogeneous graph learning tasks, owing to their superiority in capturing the intricate relationships and diverse relational semantics inherent in heterogeneous graph structures. However, the neighborhood-fetching latency incurred by structure dependency in HGNNs makes it challenging to deploy for latency-constrained applications that require fast inference. Inspired by recent GNN-to-MLP knowledge distillation frameworks, we introduce HG2M and HG2M+ to combine both HGNN's superior performance and MLP's efficient inference. HG2M directly trains student MLPs with node features as input and soft labels from teacher HGNNs as targets, and HG2M+ further distills reliable and heterogeneous semantic knowledge into student MLPs through reliable node distillation and reliable meta-path distillation. Experiments conducted on six heterogeneous graph datasets show that despite lacking structural dependencies, HG2Ms can still achieve competitive or even better performance than HGNNs and significantly outperform vanilla MLPs. Moreover, HG2Ms demonstrate a 379.24× speedup in inference over HGNNs on the large-scale IGB-3M-19 dataset, showcasing their ability for latency-sensitive deployments. Our implementation is available at: \url{https://github.com/Cloudy1225/HG2M}.
\end{abstract}


\section{Introduction}
Heterogeneous graphs are complex structures with multiple types of nodes and edges, implying complicated heterogeneous semantic relations among entities \cite{HGSurvey}. They have gained widespread popularity and application across various domains such as social networks \cite{MV-URL}, academic networks \cite{OAG}, recommender systems \cite{HERec}, knowledge graphs \cite{MRGAT}, and biological networks \cite{DeepMAPS}.

Recently, heterogeneous graph neural networks (HGNNs) have gained recognition for their ability to capture heterogeneous semantic knowledge that emerges from multiple node and edge types, as well as complex relational patterns \cite{RGCN, ieHGCN, SimpleHGN, HGAMLP}. In particular, HGNNs effectively model relational semantics, which describe the meanings of different types of connections between entities (e.g., co-actor or co-director relationships in Figure \ref{Fig: Example of HG}) \cite{HAN, MAGNN}.
HGNNs learn node representations by recursively aggregating information from neighboring nodes of various types across different relations. Through this heterogeneity-aware message-passing mechanism, they can model complex interactions, diverse entity types, and rich semantic structures, achieving state-of-the-art results in tasks such as node classification \cite{HPN}, node clustering \cite{NS4GC}, and link prediction \cite{Paths2Pair}.

\begin{figure}[ht]
    \centering
    \subfigure[\# Nodes Fetched vs. \# Layers]{\includegraphics[width=0.49\linewidth]{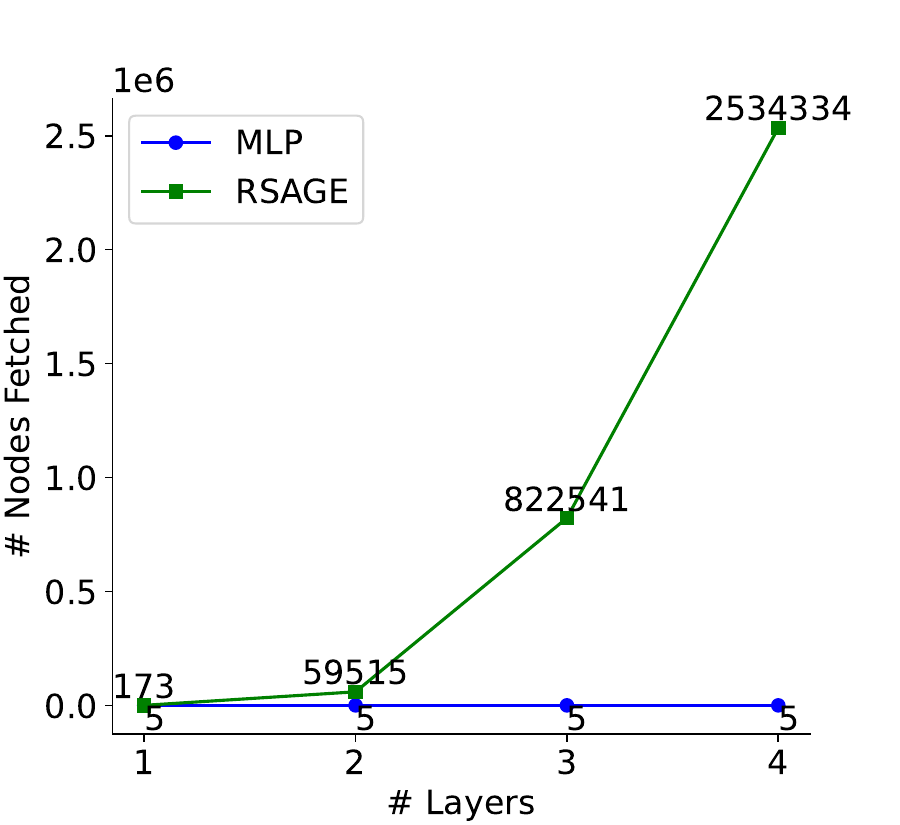}}
    \subfigure[Inference Time vs. \# Layers]{\includegraphics[width=0.49\linewidth]{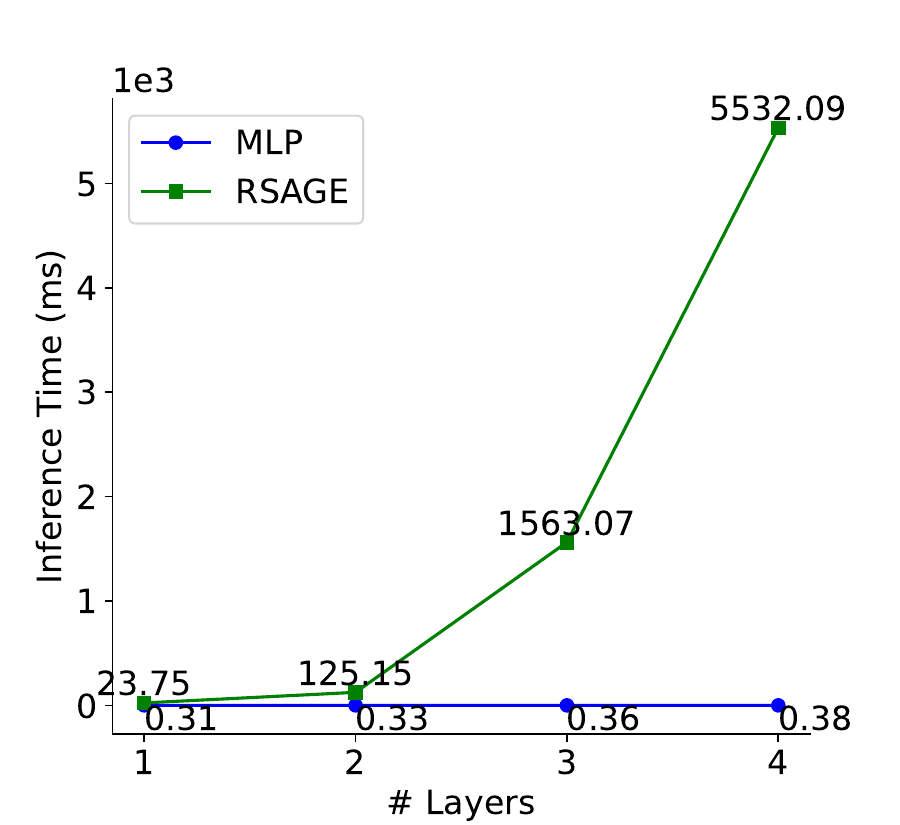}}
    \caption{The number of nodes fetched and inference time for HGNNs exceed those of MLPs by orders of magnitude, growing exponentially with the number of HGNN layers. (a) The total number of nodes fetched for inference. (b) The total inference time. (Inductive inference for $5$ random nodes on IGB-3M-19 \cite{IGB}.)}\label{Fig: Neighborhood Fetching and Inference Time}
\end{figure}

While HGNNs have achieved notable advancement, deploying them in real-world industrial settings remains challenging, especially in scenarios involving large-scale data, limited memory, and strict latency requirements. A key limitation is their reliance on the graph structure during inference. For each target node, HGNNs must fetch features from many neighbors based on the graph topology, causing both the number of fetched nodes and the inference time to grow exponentially with the number of layers \cite{GLNN}. Moreover, HGNNs process each edge type separately and often require distinct mapping functions for different node types, further increasing computational overhead. As illustrated in Figure \ref{Fig: Neighborhood Fetching and Inference Time}, deeper HGNNs result in significantly higher node dependencies and inference latency. In contrast, MLPs perform inference using only the target node's features, leading to much lower and linearly scaling inference time. However, this independence from graph structure often limits MLPs’ effectiveness on downstream tasks compared to HGNNs.


Given the trade-offs between efficiency and accuracy, several recent works \cite{GLNN, NOSMOG, KRD, VQGraph} propose GNN-to-MLP knowledge distillation frameworks where student MLPs can not only achieve comparable performance by mimicking the output of teacher GNNs but also enjoy MLPs' low-latency, dependency-free nature. However, existing research primarily focuses on homogeneous graphs, and distilling HGNNs into MLPs for heterogeneous graphs has not been explored. Heterogeneous graphs typically preserve diverse types of information that reflect complicated semantic relationships between nodes, making current GNN-to-MLP methods inadequate for capturing the heterogeneity to express the diverse semantics. Therefore, we ask: \emph{Can we bridge the gap between MLPs and HGNNs to achieve dependency-free inference and effectively distill heterogeneous semantics?}

\textbf{Present Work.} In this paper, we propose HG2M and HG2M+, which combine the superior accuracy performance of HGNNs with the efficient inference capabilities of MLPs. HG2M directly employs knowledge distillation \cite{KD} to transfer knowledge learned from teacher HGNNs to student MLPs using soft labels. Then only MLPs are deployed for fast inference, with node features as input. HG2M+ further adopts reliable node distillation and reliable meta-path distillation to inject reliable and heterogeneous semantic knowledge into student MLPs. In terms of distilling reliable knowledge, HG2M+ identifies teacher predictions with high confidence and low uncertainty as reliable knowledge nodes used for distillation. In order to distill heterogeneous semantic knowledge, HG2M+ explores the complex semantics through meta-paths \cite{HAN, MAGNN, HPN, Paths2Pair} and leverages reliable and intra-class meta-path-based neighbors to provide additional supervision for the anchor nodes.
Experiments on six real-world heterogeneous graph datasets illustrate the effectiveness and efficiency of HG2Ms. Regarding performance, under a production setting encompassing both transductive and inductive predictions, HG2M and HG2M+ exhibit average accuracy improvements of 6.24\% and 7.78\% over vanilla MLPs, respectively, and achieve competitive or even better performance to teacher HGNNs. In terms of efficiency, HG2Ms achieve an inference speedup ranging from 39.81× to 379.24× over teacher HGNNs. These results demonstrate the superior performance of HG2Ms for accurate and rapid inference in heterogeneous graph learning, making them particularly well-suited for latency-sensitive applications. To summarize, the contributions of this paper are as follows:

\begin{itemize}
    \item We are the first to integrate the superior performance of HGNNs with the efficient inference of MLPs through knowledge distillation.

    \item We design reliable node distillation and reliable meta-path distillation to inject reliable and heterogeneous semantic knowledge into student MLPs.

    \item We show HG2Ms achieve competitive performance as HGNNs and outperform vanilla MLPs significantly while enjoying 39.81×-379.24× faster inference than teacher HGNNs.
\end{itemize}

\section{Related Work}

\subsection{Heterogeneous Graph Neural Networks}
Heterogeneous Graph Neural Networks (HGNNs) use heterogeneity-aware message-passing to model intricate relationships and diverse semantics within heterogeneous graphs. RGCN \cite{RGCN} extends GCN \cite{GCN} by introducing edge type-specific graph convolutions tailored for heterogeneous graph structures. HAN \cite{HAN} employs a hierarchical attention mechanism that utilizes multiple meta-paths to aggregate node features and semantic information effectively. 
MAGNN \cite{MAGNN} encodes information from manually selected meta-paths, rather than solely endpoints. ieHGCN \cite{ieHGCN} utilizes node-level and type-level aggregation to automatically identify and exploit pertinent meta-paths for each target node, providing interpretable results. SimpleHGN \cite{SimpleHGN} incorporates a multi-layer GAT \cite{GAT} network with attention based on node features and learnable edge-type embeddings. HGAMLP \cite{HGAMLP} proposes a non-parametric framework that comprises a local multi-knowledge extractor, a de-redundancy mechanism, and a node-adaptive weight adjustment mechanism.
However, the inherent structural dependencies of HGNNs pose challenges for deploying them in latency-constrained applications requiring fast inference.

\subsection{GNN-to-MLP Knowledge Distillation}
In response to latency concerns, recent works attempt to bridge the gaps between powerful GNNs and lightweight MLPs through knowledge distillation \cite{KD}. A pioneering effort, GLNN \cite{GLNN}, directly transfers knowledge from teacher GNNs to vanilla MLPs using KL-divergence applied to their logits. To distill reliable knowledge, KRD \cite{KRD} develops a reliable sampling strategy while RKD-MLP \cite{RKD-MLP} adopts a meta-policy to filter out unreliable soft labels. FF-G2M \cite{FF-G2M} leverages both low- and high-frequency components in the spectral domain for full-frequency knowledge distillation. NOSMOG \cite{NOSMOG} introduces position features, representational similarity distillation, and adversarial feature augmentation to enhance the performance and robustness of student MLPs. VQGraph \cite{VQGraph} learns a new powerful graph representation space by directly labeling nodes' diverse local structures for GNN-to-MLP distillation. 
LLP \cite{LLP} and MUGSI further \cite{MuGSI} propose GNN-to-MLP frameworks tailored for link prediction and graph classification, while LightHGNN \cite{LightHGNN} extends this approach to hypergraphs. However, distilling HGNNs into MLPs for heterogeneous graphs has not been explored. This study aims to bridge this gap by designing a heterogeneous semantic-aware distillation approach, injecting reliable and heterogeneous semantic knowledge into MLPs to enable faster inference compared to HGNNs.

\section{Preliminary}

\begin{figure}[ht]
\centerline{\includegraphics[width=1.\linewidth]{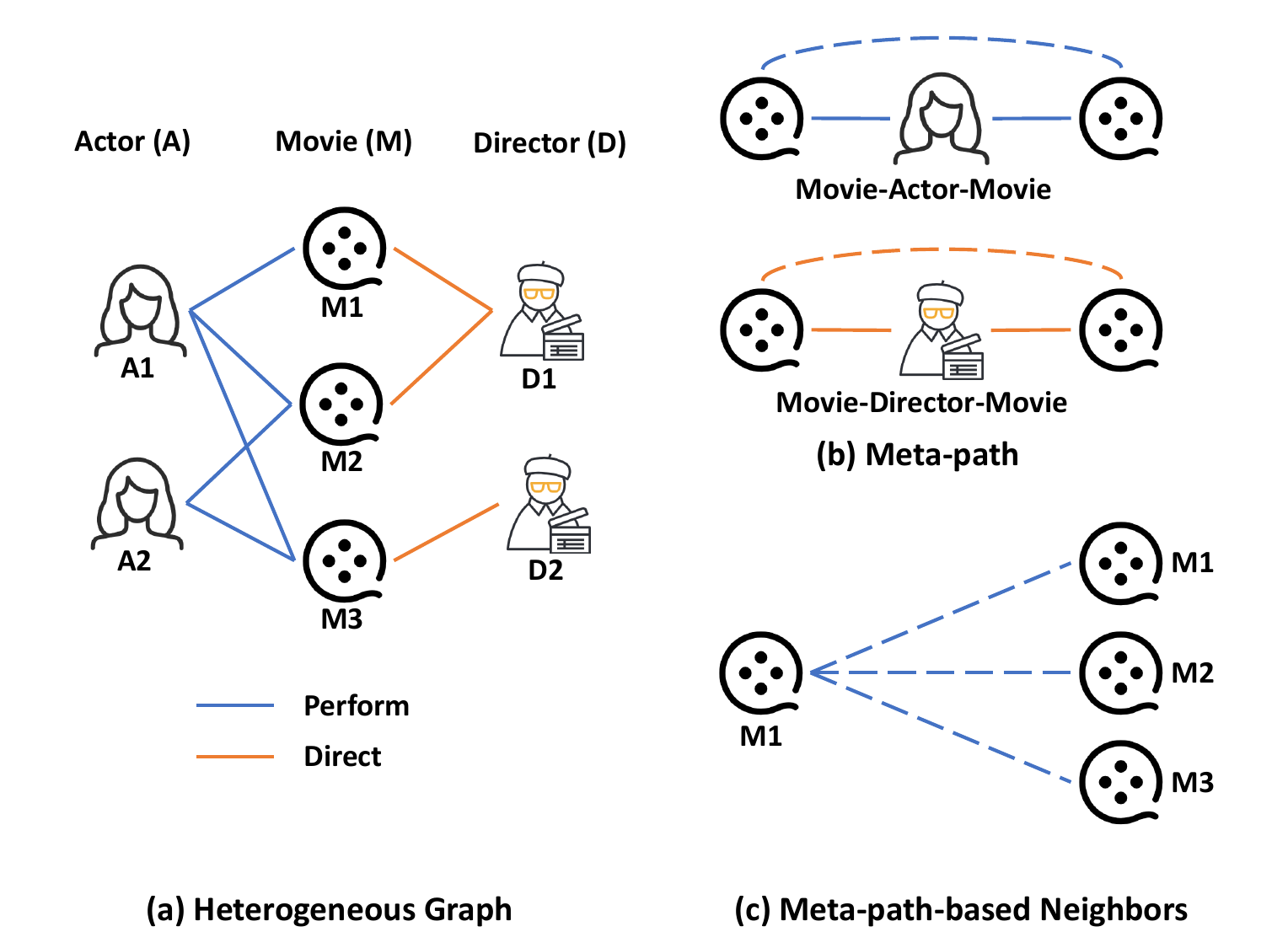}}\caption{An example of a heterogeneous graph (TMDB). 
(a) TMDB consists of three types of nodes (actor, movie, director) and two types of edges (perform, direct).
(b) TMDB involves two meta-paths: Movie-Actor-Movie and Movie-Director-Movie.
(c) Movie M1 and its meta-path-based neighbors (M1, M2, and M3).} \label{Fig: Example of HG}
\end{figure}

\subsection{Heterogeneous Graph}
\begin{definition}[Heterogeneous Graph]
A heterogeneous graph can be defined as $\mathcal{G} = ( \mathcal{V}, \mathcal{E}, \mathcal{T}, \mathcal{R}, \mathcal{X} )$, where $\mathcal{V}$ denotes the set of nodes, $\mathcal{E}$ denotes the set of edges, $\mathcal{T}$ represents the set of node types, and $\mathcal{R}$ signifies the set of edge types. 
For heterogeneous graphs, it is required that $|\mathcal{T}| + |\mathcal{R}| > 2$. 
The attribute set $\mathcal{X}=\{ \boldsymbol{X}_\tau \in \mathbb{R}^{|\mathcal{V}_\tau| \times d_\tau} \}$ includes attributes associated with each node, where $\boldsymbol{X}_\tau$ denotes the attribute matrix for a specific node type $\tau$, with $d_\tau$ representing its dimensionality.

Figure \ref{Fig: Example of HG}(a) illustrates an example heterogeneous graph with three types of nodes: actor (A), movie (M), and director (D), as well as two types of edges: perform and direct.
\end{definition}

\begin{definition}[Meta-path]
A meta-path $P$ is a composite relation among various types of nodes and edges, i.e., $P = \tau_1 \xrightarrow{r_1} \tau_2 \xrightarrow{r_2} \cdots \xrightarrow{r_l} \tau_{l+1}$ (abbreviated as $\tau_1 \tau_2 \cdots \tau_{l+1}$), where $\tau_i \in \mathcal{T}$ and $r_i \in \mathcal{R}$.

As shown in Figure \ref{Fig: Example of HG}(b), two movies can be connected via two meta-paths: Movie-Actor-Movie (MAM) and Movie-Director-Movie (MDM). Meta-path usually reveals heterogeneous semantics in a heterogeneous graph \cite{HAN, MAGNN, HPN}. For example, MAM means the co-actor relation, while MDM indicates that two movies are directed by the same director.
\end{definition}

\begin{definition}[Meta-path-based Neighbors]
Given an anchor node $v_i$ and a meta-path $P$ in a heterogeneous graph, the meta-path-based neighbors $\mathcal{N}_v^P$ of node $v$ are defined as the nodes connected to $v$ via meta-path $P$.

Referring to Figure \ref{Fig: Example of HG}(c), given the meta-path Movie-Actor-Movie, the meta-path-based neighbors of M1 are M1 (itself), M2, and M3. Similarly, the neighbors of M1 based on the meta-path Movie-Director-Movie encompass M1 and M2. These neighbors can be obtained through adjacency matrix multiplication or random walk neighborhood sampling.
\end{definition}

\begin{definition}[Node Classification]
Given a heterogeneous graph $\mathcal{G} = \{ \mathcal{V}, \mathcal{E}, \mathcal{T}, \mathcal{R}, \mathcal{X} \}$, we aim to predict the labels of the target node set $\mathcal{V}_t$ of type $t \in \mathcal{T}$. The label matrix is represented by $\boldsymbol{Y} \in \mathbb{R}^{|\mathcal{V}_t| \times k}$, where row $\boldsymbol{y}_v$ is a $k$-dimensional one-hot vector for node $v \in \mathcal{V}_t$. We use the superscript $L$ and $U$ to divide $\mathcal{V}_t$ into labeled $(\mathcal{V}_t^L, \boldsymbol{X}_t^L, \boldsymbol{Y}^L)$ and unlabeled parts $(\mathcal{V}_t^U, \boldsymbol{X}_t^U, \boldsymbol{Y}^U)$. Our objective is to predict $\boldsymbol{Y}^U$, with $\boldsymbol{Y}^L$ available. 
\end{definition}

\subsection{Heterogeneous Graph Neural Networks}
Heterogeneous Graph Neural Networks (HGNNs) utilize message-passing and aggregation techniques to integrate neighbor information across different node and edge types:
\begin{align}
    \boldsymbol{h}_{v}^{(l)} = \underset{\forall u \in \mathcal{N}(v), \forall e \in \mathcal{E}(u, v)} {\textbf{Aggregate}} \left(\textbf{Propagate}\left(\boldsymbol{h}_{u}^{(l-1)}; \boldsymbol{h}_{v}^{(l - 1)}, e\right)\right). \label{EQ: HGNN}
\end{align}
Here, $\mathcal{N}(v)$ denotes the set of source nodes connected to node $v$, and $\mathcal{E}(u, v)$ represents the edges connecting node $u$ to node $v$. In most HGNNs, the parameters of the $\textbf{Propagate}\left(\cdot\right)$ and $\textbf{Aggregate}\left(\cdot\right)$ functions depend on the types of nodes $u$ and $v$, as well as the edge $e$. This enables HGNNs to capture diverse structural semantics in the heterogeneous graph.

\section{Methodology}
In this section, we first introduce HG2M and HG2M+ to learn efficient and accurate MLPs on heterogeneous graphs. Then we provide an information-theoretical analysis of the effectiveness of HG2Ms.

\begin{figure*}
\centerline{\includegraphics[width=1.\linewidth]{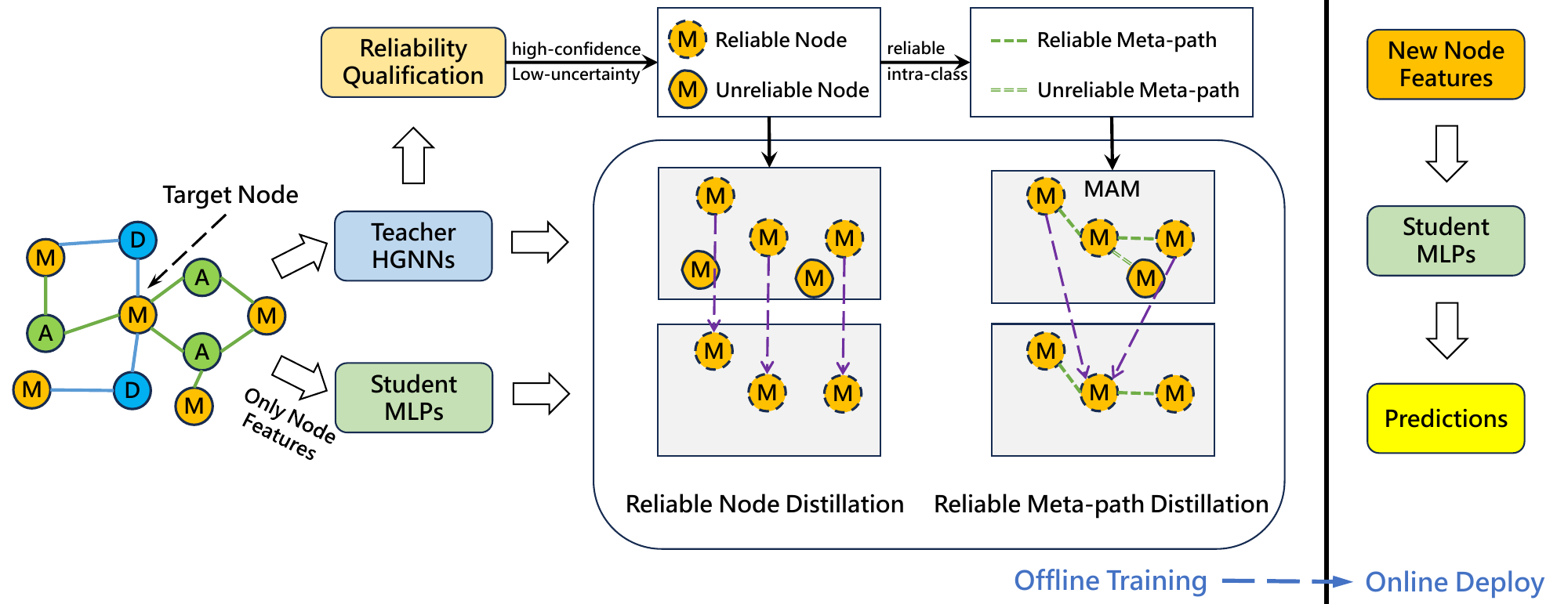}}
\caption{Overview of our proposed HG2M+. In offline training, a well-trained teacher HGNN is first used to generate soft labels on the graph. Subsequently, a student MLP is trained on node features, guided by reliable node distillation and reliable meta-path distillation. The distilled MLP, now HG2M+, is then deployed for online predictions. Since graph dependencies are eliminated during inference, HG2M+ infers much faster than HGNNs.} \label{Fig: Overview}
\end{figure*}

\subsection{HG2M}
Similar to GLNN \cite{GLNN}, the key idea of HG2M is simple yet effective: transferring knowledge from teacher HGNNs to vanilla MLPs via knowledge distillation \cite{KD}.
Specifically, we generate soft labels $\boldsymbol{z}_v$ for each target-type node $v \in \mathcal{V}_t$ using well-trained teacher HGNNs. Then we train student MLPs with both true labels $\boldsymbol{y}_v$ and $\boldsymbol{z}_v$. The objective function can be formulated as:
\begin{equation}
    \mathcal{L} = \frac{\lambda}{|\mathcal{V}_t^L|} \sum_{v \in \mathcal{V}_t^L} \mathcal{L}_{CE}\left( \hat{\boldsymbol{y}}_v, \boldsymbol{y}_v \right) + \frac{1-\lambda}{|\mathcal{V}_t|} \sum_{v \in \mathcal{V}_t} \mathcal{L}_{KL}\left( \hat{\boldsymbol{y}}_v, \boldsymbol{z}_v \right), \label{Eq: HG2M}
\end{equation}
where $\mathcal{L}_{CE}$ represents the Cross-Entropy loss between student predictions $\hat{\boldsymbol{y}}_v$ and true labels $\boldsymbol{y}_v$, $\mathcal{L}_{KL}$ denotes the Kullback-Leibler divergence loss between student predictions $\hat{\boldsymbol{y}}_v$ and soft labels $\boldsymbol{z}_v$, and $\lambda$ serves as a weight to balance these two losses. 
The model is essentially an MLP trained with cross-entropy and soft-label supervision. Therefore, during inference, HG2M has no dependency on the heterogeneous graph structure and performs as efficiently as vanilla MLPs. Additionally, through distillation, HG2M parameters are optimized to predict and generalize comparably to HGNNs, with the added benefit of faster inference and easier deployment. 

\subsection{HG2M+}
Despite the effectiveness of HG2M, it faces two critical challenges: 
(1) The soft labels of nodes incorrectly predicted by teacher HGNNs introduce noise into student MLP training, which can hinder generalization.
(2) HG2M overlooks the intricate and diverse semantic relationships embedded in heterogeneous graph structures, as it only distills node-level logits without considering relational context. 
Therefore, we further propose HG2M+, which incorporates Reliable Node Distillation (RND) to filter out noisy labels by selecting only high-confidence, low-uncertainty predictions \cite{UPS} from the teacher HGNNs, and Reliable Meta-path Distillation (RMPD) to capture higher-order heterogeneous semantics by leveraging reliable, intra-class meta-path-based neighbors \cite{HAN,MAGNN,HPN,Paths2Pair}.
These enhancements allow HG2M+ to better utilize both reliable supervision and rich heterogeneous structure, improving over the HG2M in both accuracy and robustness. The framework of HG2M+ is presented in Figure \ref{Fig: Overview}.

\subsubsection{\textbf{Reliable Node Distillation}}
As shown in Eq.~\eqref{Eq: HG2M}, the student mimics all the outputs of the teacher without selection, resulting in learning with wrong labels from unreliable predictions from the teacher model. To mitigate this problem, we introduce Reliable Node Distillation (RND), whose key idea is to filter out the wrongly predicted nodes by teacher HGNNs and construct a reliable node set $\mathcal{R}$ for training student MLPs. Formally, $\mathcal{R}=\mathcal{R}^L \cup \mathcal{R}^U$ consists of two parts, where $\mathcal{R}^L$ (or $\mathcal{R}^U$ ) includes those labeled (or unlabeled) nodes that are correctly predicted by teacher HGNNs. For labeled nodes $\mathcal{V}_t^L$, determining the reliability of a node $v \in \mathcal{V}_t^L$ is easy: if the teacher's prediction is the same as the label, the node is reliable ($v \in \mathcal{R}^L$); otherwise, it is unreliable ($v \notin \mathcal{R}^L$).


For unlabeled nodes $\mathcal{V}_t^U$, however, assessing the correctness of the teacher's predictions poses a challenge due to the absence of known labels. To address this, we adopt a widely used pseudo-labeling assumption: a predicted label is likely to be the true label if the model predicts it with high confidence and low uncertainty \cite{UPS, RDD, CPL}. Given an unlabelled node $v \in \mathcal{V}_t^U$ and its soft labels $\boldsymbol{z}_v \in [0,1]^{k}$ predicted by teacher HGNNs, the prediction confidence can be defined as the probability of the predicted label $\max(\boldsymbol{z}_v)$, and the prediction uncertainty can be measured by the information entropy $\mathcal{H}(\boldsymbol{z}_v) = -\sum_i z_{v, i} \log(z_{v, i})$. Then, the set of reliable unlabeled nodes can be expressed as:
\begin{equation}
    \mathcal{R}^U = \left \{ v | \max(\boldsymbol{z}_v) \geq \tau_c \wedge \mathcal{H}(\boldsymbol{z}_v) \leq \tau_u, v \in \mathcal{V}_t^U \right \}, 
\end{equation}
where $\tau_c$ and $\tau_u$ are the confidence and uncertainty thresholds, respectively. A node is deemed reliable ($v \in \mathcal{R}^U$) if both the prediction confidence is sufficiently high ($\max(\boldsymbol{z}_v) \geq \tau_c$) and the prediction uncertainty is sufficiently low ($\mathcal{H}(\boldsymbol{z}_v) \leq \tau_u$); otherwise, it is considered unreliable ($v \notin \mathcal{R}^U$). Notably, instead of using a threshold that may vary significantly for different data and models, we identify the $p$-percent of predictions with the highest confidence and lowest uncertainty as reliable, with all others considered unreliable. 

During distillation, only reliable nodes are utilized, while unreliable nodes are discarded. Thus, the loss function for Reliable Node Distillation is formulated as:
\begin{equation}
    \mathcal{L}_{RND} = \frac{1}{|\mathcal{R}|} \sum_{v \in \mathcal{R}} \mathcal{L}_{KL}\left( \hat{\boldsymbol{y}}_v, \boldsymbol{z}_v \right).
\end{equation}

\subsubsection{\textbf{Reliable Meta-Path Distillation}}
Existing GNN-to-MLP methods and HG2M ignore the intricate relationships and diverse relational semantics inherent in heterogeneous graph structures, resulting in suboptimal performance. Meta-paths typically capture complex semantics and higher-order structures in heterogeneous graphs \cite{HAN, MAGNN, HPN, Paths2Pair}. Therefore, we propose Reliable Meta-Path Distillation (RMPD) to achieve heterogeneous semantic-aware distillation. The key idea of RMPD is to transfer valuable knowledge from meta-path-based neighbors to an anchor node. Crucially, these selected meta-path-based neighbors must be both \emph{reliable} and \emph{intra-class} (i.e., share the same label as the anchor node), since unreliable and inter-class neighbors could introduce misleading information to the anchor node. 

Given a meta-path $P$ with the node type of endpoints as the target type $t$, we can construct the $P$-induced meta-path subgraph $\mathcal{G}^P = ( \mathcal{V}_t,  \mathcal{E}^P )$, where $\mathcal{E}^P$ comprises all the meta-path-$P$-based neighbor pairs. One critical step of RMPD is to select reliable and intra-class meta-path-$P$-based neighbor pairs $\mathcal{M}_P$ from $\mathcal{E}^P$. Similar to RND, $\mathcal{M}_P = \mathcal{M}_P^L \cup \mathcal{M}_P^U$, where $\mathcal{M}_P^L$ ($\mathcal{M}_P^U$) requires pairs with both endpoints labeled (at least one endpoint unlabeled). We now describe how to construct $\mathcal{M}_P$. For clarity, let $v$ denote the anchor node and $u$ denote one of its meta-path-$P$-based neighbors $\mathcal{N}_v^P$. For meta-path-$P$-based neighbor pairs with both endpoints labeled, i.e., $u, v \in \mathcal{V}_t^L$, if $u$ is correctly predicted by teacher HGNNs (\emph{reliable}) and shares the same label as $v$ (\emph{intra-class}), this meta-path-$P$-based neighbor pair is reliable, i.e., $(u,v) \in \mathcal{M}_P^L$; otherwise, it is unreliable, i.e., $(u,v) \notin \mathcal{M}_P^L$.

For meta-path-$P$-based neighbor pairs where at least one endpoint is unlabeled, identifying reliable neighbors is facilitated by the pre-established reliable node set $\mathcal{R}$ in RND. However, determining whether $u$ and $v$ share the same label remains a challenge. To tackle this issue, we approach it as a binary classification task aimed at estimating the probability that $u$ and $v$ belong to the same class. Specifically, we first extract all meta-path-$P$-based neighbor pairs from $\mathcal{E}^P$ where the neighbor $u$ is deemed \emph{reliable} ($u \in \mathcal{R}$). These pairs are then split into training and test sets based on whether nodes $u$ and $v$ are labeled. Each pair is labeled binary according to whether $u$ and $v$ share the same class: a positive label is assigned if $u$ belongs to the same class as the anchor node $v$; otherwise, it receives a negative label. Assuming that nodes $u$ and $v$ are more likely to share the same class if they exhibit similar attributes, predicted soft labels, and are connected through multiple meta-path instances \cite{HeCo, OGB-LSC}, we construct a 3-dimensional feature vector $\boldsymbol{f}_{uv} = \left[ f_{uv}^1, f_{uv}^2, f_{uv}^3 \right]$ for each pair $(u,v)$, capturing attribute similarity, structural connections, and label similarity. 
Here, $f_{uv}^1 = \frac{\boldsymbol{x}_u \cdot \boldsymbol{x}_v}{\parallel \boldsymbol{x}_u \parallel \parallel \boldsymbol{x}_v \parallel}$ represents the cosine similarity between the attribute vectors $\boldsymbol{x}_u$ and $\boldsymbol{x}_v$; $f_{uv}^2$ denotes the connection strength between $u$ and $v$, i.e., the count of different possible paths along meta-path $P$; and $f_{uv}^3 = \boldsymbol{z}_u \cdot \boldsymbol{z}_v$ signifies the inner product of the predicted soft label vectors $\boldsymbol{z}_u$ and $\boldsymbol{z}_v$. Finally, we employ a Logistic Regression (LR) classifier with an L2 penalty well-trained on the training data to predict the probability $p_{uv}$ that each meta-path-$P$-based pair $(u,v)$ in the test set belongs to the same class. If $p_{uv} > 0.5$, $(u,v)$ is \emph{intra-class} and $(u,v) \in \mathcal{M}_P^U$; otherwise, $(u,v)$ is inter-class and $(u,v) \notin \mathcal{M}_P^U$.

Given a set of meta-paths $\mathcal{P}$, we obtain the reliable and intra-class meta-path-based neighbor pair set $\mathcal{M}_P$ for each meta-path $P \in \mathcal{P}$. Then we transfer valuable knowledge from these meta-path-based neighbors to the anchor node using the following loss:
\begin{equation}
    \mathcal{L}_{RMPD} = \frac{1}{|\mathcal{P}|}\sum_{P \in \mathcal{P}}\frac{1}{|\mathcal{M}_P|}\sum_{(u,v) \in \mathcal{M}_P} p_{uv} \mathcal{L}_{KL}(\hat{\boldsymbol{y}}_v, \boldsymbol{z}_u),
\end{equation}
where $p_{uv} = 1.0$ for $(u,v)$ belongs to the labeled set $\mathcal{M}_P^L$. Meta-paths are known for capturing intricate semantics and higher-order structures in heterogeneous graphs \cite{HAN, MAGNN, HPN, Paths2Pair}. Consequently, RMPD further enhances HG2M by distilling heterogeneous semantic knowledge, as shown in Section \ref{Sec: Ablation Study}.

With reliable node distillation and reliable meta-path distillation, the overall loss function for HG2M+ is defined as follows:


\begin{equation}
\begin{aligned}
    \mathcal{L} 
    &= \frac{\lambda}{|\mathcal{V}_t^L|} \sum_{v \in \mathcal{V}_t^L} \mathcal{L}_{CE}\left( \hat{\boldsymbol{y}}_v, \boldsymbol{y}_v \right) + 
    (1-\lambda) ( \frac{1}{|\mathcal{R}|} \sum_{v \in \mathcal{R}} \mathcal{L}_{KL}\left( \hat{\boldsymbol{y}}_v, \boldsymbol{z}_v \right)\\ 
    &+ 
    \frac{1}{|\mathcal{P}|}\sum_{P \in \mathcal{P}}\frac{1}{|\mathcal{M}_P|}\sum_{(u,v) \in \mathcal{M}_P} p_{uv} \mathcal{L}_{KL}(\hat{\boldsymbol{y}}_v, \boldsymbol{z}_u) \Large). \label{Eq: HG2M+}
\end{aligned}
\end{equation}

\subsection{Why do HG2Ms work?}\label{Sec: Theoretical Analysis}
Intuitively, incorporating neighbor information makes HGNNs more powerful than MLPs for node classification. However, as shown in Section \ref{Sec: Performance Comparison}, with knowledge distillation, HG2Ms can achieve competitive or even superior performance to teacher HGNNs. To provide more insights, we analyze the effectiveness of HG2Ms from an information-theoretic perspective.

The goal of node classification is to learn a function $f$ on the rooted graph $\mathcal{G}^{[v]}$ with label $\boldsymbol{y}_v$ \cite{GAMLP}. From an information-theoretic perspective, learning $f$ by minimizing cross-entropy loss is equivalent to maximizing the mutual information $I(\mathcal{G}^{[v]}; \boldsymbol{y}_i)$ \cite{qin2019rethinking}. If we treat $\mathcal{G}^{[v]}$ as the joint distribution of two random variables $\boldsymbol{X}^{[v]}$ (node features) and $\mathcal{E}^{[v]}$ (edges), the mutual information can be expressed as:
\begin{equation}
    I(\mathcal{G}^{[v]}; \boldsymbol{y}_v) = I(\boldsymbol{X}^{[v]}, \mathcal{E}^{[v]}; \boldsymbol{y}_v) =  I(\mathcal{E}^{[v]}; \boldsymbol{y}_v) + I(\boldsymbol{X}^{[v]}; \boldsymbol{y}_v \vert \mathcal{E}^{[v]}),
\end{equation}
where $I(\mathcal{E}^{[v]}; \boldsymbol{y}_v)$ is the mutual information between edges and labels, indicating the relevance of the graph structure to labels, while $I(\boldsymbol{X}^{[v]}; \boldsymbol{y}_v \vert \mathcal{E}^{[v]})$ is the mutual information between features and labels conditioned on edges $\mathcal{E}^{[v]}$. 
HG2M leverages the objective function defined in Eq.~\eqref{Eq: HG2M}, which approximates $I(\mathcal{G}^{[v]}; \boldsymbol{y}_v)$ by only maxmizing $I(\boldsymbol{X}^{[v]}; \boldsymbol{y}_v \vert \mathcal{E}^{[v]})$ while ignoring $I(\mathcal{E}^{[v]}; \boldsymbol{y}_v)$. But for real-world node classification tasks, node features and structural roles are often highly correlated \cite{AN2VEC, GLNN}, allowing MLPs to perform reasonably well even when relying solely on node features. Then with the help of knowledge distillation, MLP parameters approximate the ideal prediction function from node features to labels, thus HG2M can potentially achieve better results. Additionally, to leverage the complementary structural information, HG2M+ uses meta-path structures to model $I(\mathcal{E}^{[v]};\boldsymbol{y}_v)$, so it can work better in the case that labels are also correlated to the graph structure. Even in the extreme case where $\boldsymbol{y}_v$ is uncorrelated with $I(\boldsymbol{X}^{[v]}; \boldsymbol{y}_v \vert \mathcal{E}^{[v]})$, HG2M+ can still achieve superior or comparable performance to HGNNs, as shown in Section \ref{Sec: Noisy Node Features}.

\begin{algorithm}[ht]
\caption{Algorithm of HG2M+} \label{Alg: HG2M+}
\begin{algorithmic}[1]
\STATE {\bfseries Input:} The heterogeneous graph $\mathcal{G} = ( \mathcal{V}, \mathcal{E}, \mathcal{T}, \mathcal{R}, \mathcal{X} )$, labels $\boldsymbol{Y}^L$, \# epochs $E$, meta-paths $\mathcal{P}$.
\STATE {\bfseries Output:} Predicted labels $\hat{\boldsymbol{Y}}^U$ for unlabeled nodes, student MLPs' parameters $\mathbf{\Theta}$.
\STATE Pre-train teacher HGNNs with labels $\boldsymbol{Y}^L$.
\STATE Select reliable nodes $\mathcal{R}$ in RND.
\STATE Select reliable and intra-class meta-path-based neighbor pairs $\mathcal{M}_P$ for each $P \in \mathcal{P}$ in RMPD.
\STATE Initialize the parameters $\mathbf{\Theta}$ of student MLPs.
\FOR{$e = 1$ to $E$}
\STATE Calculate the total loss of cross-entropy loss, RND loss, and RMPD loss by Eq.~\eqref{Eq: HG2M+}.
\STATE Update student MLPs' parameters $\mathbf{\Theta}$ by back propagation.
\ENDFOR
\STATE {\bfseries Return} Predicted labels $\hat{\boldsymbol{Y}}^U$ for unlabeled vertices, student MLPs' parameters $\mathbf{\Theta}$.
\end{algorithmic}
\end{algorithm}

\section{Experiments}
In this section, we conduct a series of experiments to evaluate the effectiveness and efficiency of the proposed HG2Ms by investigating the following research questions:
\begin{description}
    \item[RQ1:] How do HG2Ms compare to MLPs and HGNNs under both transductive and inductive settings?

    \item[RQ2:] How do HG2Ms compare to other GNN-to-MLP knowledge distillation methods?
    
    \item[RQ3:] How efficient are HG2Ms compared to HGNNs?

    \item[RQ4:] How does each component contribute to HG2M+?
                
    \item[RQ5:] How does each meta-path contribute to HG2M+?
                
    \item[RQ6:] How do HG2Ms perform with different teachers?

    \item[RQ7:] How do HG2Ms perform with noisy node features?

    \item[RQ8:] How do different hyperparameters affect HG2Ms?
\end{description}

\subsection{Experimental Setup}

\begin{table}[ht]
\centering
\caption{Statistics of datasets. \textbf{Bold} numbers are the total count of nodes or edges, while \underline{underlined} node types are the target nodes for classification.}\label{Tab: Dataset Statistics}
\resizebox{0.46\textwidth}{!}{\begin{tabular}{c|l|l|c|c} 
\toprule
Dataset               & \# Nodes       & \# Edges    & Meta-paths & \# Classes          \\ 
\midrule
\multirow{4}{*}{TMDB}         & \textbf{24,412}               & \textbf{104,858} &        & \multirow{4}{*}{4}  \\
                              & \underline{Movie}: 7,505      & M-A: 86,517      & MAM    &                     \\
                              & Actor: 13,016                 & M-D: 18,341      & MDM    &                     \\
                              & Director: 3,891               &                  &        &                     \\
\midrule
\multirow{4}{*}{CroVal}       & \textbf{44,386}               & \textbf{118,712} &        & \multirow{4}{*}{6}  \\
                              & \underline{Question}: 34,153  & Q-U: 34,153      & QUQ    &                     \\
                              & User: 8,898                   & Q-T: 84,559      & QTQ    &                     \\
                              & Tag: 1,335                    &                  &        &                     \\
\midrule
\multirow{3}{*}{ArXiv}        & \textbf{209,224}              & \textbf{841,839} &        & \multirow{3}{*}{40} \\
                              & \underline{Paper}: 81,634     & P-P: 541,606     & PP     &  \\
                              & Author: 127,590               & P-A: 300,233     & PAP    &                     \\
\midrule
\multirow{5}{*}{IGB-549K-19}  & \textbf{549,999}              & \textbf{2,046,541} &       & \multirow{5}{*}{19} \\
                              & \underline{Paper}: 100,000    & P-P: 547,076   & PP        &                     \\
                              & Author: 357,041               & P-A: 455,610   & PAP       &                     \\
                              & Institute: 8,738              & A-I: 325,410   &           &                     \\
                              & FoS: 84,220                   & P-F: 718,445   &           &                     \\ 
\midrule
\multirow{5}{*}{IGB-549K-2K}  & \textbf{549,999}              & \textbf{2,046,541} &       & \multirow{5}{*}{2,983} \\
                              & \underline{Paper}: 100,000    & P-P: 547,076   & PP        &                     \\
                              & Author: 357,041               & P-A: 455,610   & PAP       &                     \\
                              & Institute: 8,738              & A-I: 325,410   &           &                     \\
                              & FoS: 84,220                   & P-F: 718,445   &           &                     \\ 
\midrule
\multirow{5}{*}{IGB-3M-19}    & \textbf{3,131,266}            & \textbf{26,334,780} &      & \multirow{5}{*}{19} \\
                              & \underline{Paper}: 1,000,000  & P-P: 13,068,130& PP        &                     \\
                              & Author: 1,926,066             & P-A: 4,402,052 & PAP       &                     \\
                              & Institute: 14,751             & A-I: 1,630,476 &           &                     \\
                              & FoS: 190,449                  & P-F: 7,234,122 &           &                     \\ 
\bottomrule
\end{tabular}}
\end{table}

\begin{table*}[ht]
    \begin{center}
    \begin{threeparttable}
    \caption{{Hyper-parameters for MLPs, HGNNs, and HG2Ms.}\label{Tab: Hyperparameters}}
    \begin{tabular}{lccccccc}
    \toprule
    ~   & MLP & RSAGE & RGCN & RGAT & SimpleHGN & ieHGCN & HG2Ms \\ \midrule
    \# layers$^1$ & 2 & 2 & 2 & 2 & 2 & 2 & 2 \\
    hidden dim$^1$ & 128 & 128 & 128 & 128 & 128 & 128 & 128 \\
    learning rate & 0.01 & 0.01 & 0.01 & 0.01 & 0.01 & 0.01 & 0.01 \\
    weight decay & 0.0005 & 0 & 0 & 0 & 0 & 0 & 0 \\
    dropout & 0.2 & 0.2 & 0.2 & 0.2 & 0.5 & 0.5 & 0.2 \\
    fan out$^1$ & - & [10, 15] & [10, 15] & [10, 15] & - & - & - \\
    \# attention heads & - & - & - & 4 & 4 & -  & -\\
    \bottomrule
    \end{tabular}
    \begin{tablenotes}
        \item[1] On the AriXiv dataset, the \# layers, the fan out for HGNNs, and the hidden dim for MLPs/HG2Ms are set to 3, [10, 15, 15], and 512 respectively.
    \end{tablenotes}
    \end{threeparttable}
    \end{center}
\end{table*}

\subsubsection{Datasets}
We evaluate HG2Ms on six real-world datasets, including three small datasets TMDB, CroVal, ArXiv \cite{HTAG}  and three more larger IGB datasets \cite{IGB}: IGB-549K-19, IGB-549K-2K, and IGB-3M-19. In Table \ref{Tab: Dataset Statistics}, we present the statistical information of the datasets used in our experiments. Further details about each dataset are provided below.

\begin{itemize}
\item \textbf{TMDB}\footnote{\url{https://www.themoviedb.org}} is a popular online database and community platform that provides a vast collection of information about movies, TV shows, and other related content. We use a subset of TMDB obtained via the platform's public API\footnote{\url{https://developer.themoviedb.org/docs}} as of May 31, 2024. It contains 7,505 of the most popular movies, 13,016 actors, and 3,891 directors. Movies are labeled as one of 4 classes (\textit{Action}, \textit{Romance}, \textit{Thriller}, and \textit{Animation}) based on their genre. We pass the movie overview to a MiniLM\footnote{\url{https://huggingface.co/sentence-transformers/all-MiniLM-L6-v2}} \cite{MiniLM} sentence encoder \cite{Sentence-BERT}, generating a 384-dimensional feature vector for each movie node. We train on movies released up to 2015, validate on those released from 2016 to 2018, and test on those released since 2019.

\item \textbf{CroVal}\footnote{\url{https://stats.stackexchange.com}} is a question-and-answer website for people interested in statistics, machine learning, data analysis, data mining, and data visualization. We use the version of Cross Validated data dump\footnote{\url{https://archive.org/download/stackexchange}} released on April 6, 2024. After data preprocessing, it contains 34,153 questions, 8,893 users, and 1,335 tags. Questions are categorized into six classes (\textit{Regression}, \textit{Hypothesis-Testing}, \textit{Distributions}, \textit{Neural-Networks}, \textit{Classification}, \textit{Clustering}) based on their topic. We pass the question title to a MiniLM sentence encoder, generating a 384-dimensional feature vector for each question node. We train on questions posted up to 2015, validate on those posted from 2016 to 2018, and test on those posted since 2019.

\item \textbf{ArXiv}\footnote{\url{https://arxiv.org/search/cs}} is an academic network dataset, containing 81,634 Computer Science (CS) arXiv papers indexed by Microsoft Academic Graph and 127,590 authors. Papers are categorized into 40 subject areas of arXiv CS papers (e.g., cs.AI, cs.LG, and cs.OS). Each paper comes with a 128-dimensional feature vector derived from the averaged word2vec embeddings of its title and abstract. We train on papers published until 2017, validate on those published in 2018, and test on those published since 2019.

\item \textbf{IGB} \cite{IGB} is a heterogeneous academic graph constructed from Microsoft Academic Graph. It contains four types of nodes: papers, authors, institutes, and fields of study (FoS). Papers are annotated with two different numbers of classes (19 and 2,983) based on the granularity of paper topics. Each paper is associated with a 1024-dimensional feature vector generated by passing its title and abstract to a RoBERTa \cite{RoBERTa} sentence encoder. We use three datasets from IGB of varying sizes: IGB-549K-19, IGB-549K-2K, and IGB-3M-19. We train on papers published until 2016, validate on those published in 2017 and 2018, and test on those published since 2019.
\end{itemize}

We split all datasets according to \emph{time}, which is more realistic and challenging \cite{OGB-LSC}. For node types lacking initial features, we derive these features from nodes that do possess them. For example, in the ArXiv and IGB datasets, author node features are computed by averaging the features of all papers written by that author \cite{OGB-LSC, IGB}.

\subsubsection{Model Architectures} For a fair comparison, we use RSAGE, a GraphSAGE \cite{GraphSAGE} extension to heterogeneous graphs, with GCN \cite{GCN} aggregation as the teacher. We also investigate the influence of alternative teacher models, including RGCN \cite{RGCN}, RGAT \cite{RGAT}, SimpleHGN \cite{SimpleHGN}, and ieHGCN \cite{ieHGCN}, as detailed in Section \ref{Sec: Model Architecture}.

\subsubsection{Evaluation Protocol}
We report the average and standard deviation over five runs with different random seeds. We adopt accuracy to measure the model performance and select the model with the highest validation accuracy for testing.

\subsubsection{Implementation Details}
The experiments on both baselines and our approach are implemented using PyTorch and DGL for GNN algorithms, and Adam \cite{Adam} for optimization. We run all experiments on a 32GB NVIDIA Tesla V100 GPU. The hyperparameters for vanilla MLPs, teacher HGNNs, and HG2Ms are detailed in Table \ref{Tab: Hyperparameters}. HG2Ms are configured with the same number of layers and hidden dimensions per layer as teacher HGNNs to maintain the same parameter counts. The parameter $\lambda$ in Eq.~\eqref{Eq: HG2M} and Eq.~\eqref{Eq: HG2M+} balances the weight of supervision from the true labels and soft labels. In our experiments, we performed slight tuning of $\lambda$ but found that non-zero values did not significantly improve results. Therefore, following the approach in \cite{GLNN, LightHGNN, MGFNN}, we present results with $\lambda = 0$, where only the distillation term is effective. Additionally, based on empirical findings in Section \ref{Sec: Reliable Node Proportion}, we set the reliable node proportion $p = 0.9$ consistently across all datasets. For RMPD in HG2M+, we use the meta-paths listed in Table \ref{Tab: Dataset Statistics}. To ensure reproducibility, our implementation is publicly available at \url{https://github.com/Cloudy1225/HG2M}.

\subsubsection{Transductive vs. Inductive}\label{Sec: Transductive vs. Inductive}
To comprehensively evaluate our model, we perform node classification under two settings: transductive (\textit{tran}) and inductive (\textit{ind}). In the \textit{tran} setting, we train models on $\mathcal{G}$, $\boldsymbol{X}_t^L$, and $\boldsymbol{Y}^L$, while evaluating them on $\boldsymbol{X}_t^U$ and $\boldsymbol{Y}^U$. During distillation, the entire graph including the validation and test nodes is used to generate soft labels $\boldsymbol{z}_v$ for every target-type node $v \in \mathcal{V}_t$. For the \textit{ind} setting, we follow GLNN \cite{GLNN} to randomly select out 20\% test data for inductive evaluation. Specifically, we divide the unlabeled nodes $\mathcal{V}_t^U$ into two disjoint subsets: observed $\mathcal{V}_{t, obs}^U$ and inductive $\mathcal{V}_{t, ind}^U$, resulting in three distinct graphs $\mathcal{G} = \mathcal{G}^L \cup \mathcal{G}_{obs}^U \cup \mathcal{G}_{ind}^U$ with no shared target-type nodes. During training, edges between $\mathcal{G}^L \cup \mathcal{G}_{obs}^U$ and $\mathcal{G}_{ind}^U$, as well as non-target-type nodes only in $\mathcal{G}_{ind}^U$, are removed, but they are utilized during inference. Target-type node and labels are partitioned into three disjoint sets: $\boldsymbol{X}_t^L \cup \boldsymbol{X}_{t, obs}^U \cup \boldsymbol{X}_{t, ind}^U$ and $\boldsymbol{Y}^L \cup \boldsymbol{Y}_{obs}^U \cup \boldsymbol{Y}_{ind}^U$. Soft labels are generated during distillation for nodes in the labeled and observed subsets, i.e., $\boldsymbol{z}_v$ for $v \in \mathcal{V}_t^L \cup \mathcal{V}_{t, obs}^U$.

\begin{table*}[!ht]
    \begin{center}
    {\caption{Classificatiom accuracy ± std (\%) under the transductive setting. $\Delta_{MLP}$, $\Delta_{HGNN}$, $\Delta_{HG2M}$ represents the difference between the HG2Ms and MLP, RSAGE, HG2M, respectively.}\label{Tab: Transductive Setting}}
    \begin{tabular}{c|cc|ccc|cccc}
    \toprule
    Dataset & MLP & RSAGE & HG2M & $\Delta_{MLP}$ & $\Delta_{HGNN}$ & HG2M+ & $\Delta_{MLP}$ & $\Delta_{HGNN}$ & $\Delta_{HG2M}$ \\ \midrule
    TMDB & 72.19±0.34 & 81.11±0.08 & 80.64±0.44 & 8.45 & -0.47 & \textbf{84.69±0.39} & 12.50 & 3.58 & 4.05 \\
    CroVal & 82.35±0.24 & 86.56±0.12 & 86.16±0.20 & 3.81 & -0.40 & \textbf{89.05±0.10} & 6.70 & 2.49 & 2.89 \\
    ArXiv & 64.36±0.20 & 76.11±0.11 & 76.46±0.10 & 12.10 & 0.35 & \textbf{78.80±0.30} & 14.44 & 2.69 & 2.34 \\
    IGB-549K-19 & 55.81±0.17 & 58.53±0.55 & 57.99±0.49 & 2.18 & -0.54 & \textbf{59.62±0.49} & 3.81 & 1.09 & 1.63 \\
    IGB-549K-2K & 53.20±0.14 & 58.06±0.47 & 57.34±0.50 & 4.14 & -0.72 & \textbf{59.20±0.66} & 6.00 & 1.14 & 1.86 \\
    IGB-3M-19 & 62.35±0.15 & \textbf{66.36±0.67} & 64.89±0.40 & 2.54 & -1.47 & 66.27±0.40 & 3.92 & -0.09 & 1.38 \\ 
    \midrule
    Avg. Rank/Avg. & 4.0 & 2.0 & 2.8 & 5.54 & -0.54 & 1.2 & 7.90 & 1.82 & 2.36 \\
    \bottomrule
    \end{tabular}
    \end{center}
\end{table*}

\begin{table*}[!ht]
    \begin{center}
    {\caption{Classificatiom accuracy ± std (\%) under the production setting with both inductive and transductive predictions. \textit{ind} results on $\mathcal{V}_{t, ind}^U$, \textit{tran} results on $\mathcal{V}_{t, obs}^U$, and the interpolated \textit{prod} results are reported ($prod=0.2\times ind+0.8\times tran$).}\label{Tab: Production Setting}}
    \begin{tabular}{cc|cc|ccc|cccc}
    \toprule
    Dataset & Eval & MLP & RSAGE & HG2M & $\Delta_{MLP}$ & $\Delta_{HGNN}$ & HG2M+ & $\Delta_{MLP}$ & $\Delta_{HGNN}$ & $\Delta_{HG2M}$ \\ \midrule
    \multirow{3}{*}{TMDB} & \textit{prod} & 72.17±0.28 & \textbf{81.97±0.55} & 79.58±0.52 & 7.41 & -2.39 & 81.11±0.68 & 8.94 & -0.86 & 1.53 \\
    ~ & \textit{ind} & 70.68±2.85 & 80.09±2.28 & 71.60±2.03 & 0.92 & -8.49 & 72.24±2.44 & 1.56 & -7.85 & 0.64 \\
    ~ & \textit{tran} & 72.54±0.51 & 82.44±0.37 & 81.57±0.46 & 9.03 & -0.87 & 83.33±0.37 & 10.79 & 0.89 & 1.76 \\ \midrule
    \multirow{3}{*}{CroVal} & \textit{prod} & 82.45±0.12 & 86.32±0.15 & 85.55±0.19 & 3.10 & -0.77 & \textbf{88.02±0.16} & 5.57 & 1.70 & 2.47 \\
    ~ & \textit{ind} & 82.36±0.68 & 86.06±0.76 & 83.14±0.29 & 0.78 & -2.92 & 83.76±0.44 & 1.40 & -2.30 & 0.62 \\
    ~ & \textit{tran} & 82.47±0.08 & 86.39±0.24 & 86.15±0.24 & 3.68 & -0.24 & 89.09±0.22 & 6.62 & 2.70 & 2.94 \\ \midrule
    \multirow{3}{*}{ArXiv} & \textit{prod} & 64.25±0.16 & \textbf{70.76±0.16} & 68.91±0.17 & 4.66 & -1.85 & 70.34±0.30 & 6.09 & -0.42 & 1.43 \\
    ~ & \textit{ind} & 64.02±0.97 & 77.55±1.00 & 65.81±0.48 & 1.79 & -11.74 & 65.95±0.77 & 1.93 & -11.60 & 0.14 \\
    ~ & \textit{tran} & 64.30±0.25 & 69.07±0.21 & 69.69±0.30 & 5.39 & 0.62 & 71.44±0.29 & 7.14 & 2.37 & 1.75 \\ \midrule
    \multirow{3}{*}{IGB-549K-19} & \textit{prod} & 55.83±0.16 & 67.42±0.41 & 67.53±0.18 & 11.70 & 0.11 & \textbf{69.15±0.31} & 13.32 & 1.73 & 1.62 \\
    ~ & \textit{ind} & 55.19±1.94 & 57.88±1.56 & 58.16±1.63 & 2.97 & 0.28 & 57.85±1.64 & 2.66 & -0.03 & -0.31 \\
    ~ & \textit{tran} & 55.99±0.32 & 69.81±0.24 & 69.88±0.25 & 13.89 & 0.07 & 71.97±0.45 & 15.98 & 2.16 & 2.09 \\ \midrule
    \multirow{3}{*}{IGB-549K-2K} & \textit{prod} & 53.34±0.13 & 59.53±0.41 & 58.99±0.63 & 5.65 & -0.54 & \textbf{60.79±0.69} & 7.45 & 1.26 & 1.80 \\
    ~ & \textit{ind} & 52.84±1.79 & 54.25±1.48 & 57.16±1.63 & 4.32 & 2.91 & 57.65±2.17 & 4.81 & 3.40 & 0.49 \\
    ~ & \textit{tran} & 53.47±0.56 & 60.86±0.16 & 59.44±0.42 & 5.97 & -1.42 & 61.58±0.47 & 8.11 & 0.72 & 2.14 \\ \midrule
    \multirow{3}{*}{IGB-3M-19} & \textit{prod} & 62.30±0.08 & \textbf{69.02±0.19} & 67.21±0.21 & 4.91 & -1.81 & 67.62±0.29 & 5.32 & -1.40 & 0.41 \\
    ~ & \textit{ind} & 61.83±1.14 & 65.12±1.08 & 64.11±1.02 & 2.28 & -1.01 & 64.04±1.23 & 2.21 & -1.08 & -0.07 \\
    ~ & \textit{tran} & 62.42±0.35 & 70.00±0.26 & 67.99±0.13 & 5.57 & -2.01 & 68.52±0.08 & 6.10 & -1.48 & 0.53 \\
    \midrule
    \multicolumn{2}{c|}{Avg. Rank/Avg.} & 4.0 & 1.7 & 2.8 & 6.24 & -1.21 & 1.5 & 7.78 & 0.34 & 1.54 \\
    \bottomrule
    \end{tabular}
    \end{center}
\end{table*}

\subsection{How do HG2Ms compare to MLPs and HGNNs? (RQ1)}\label{Sec: Performance Comparison}
We begin by comparing HG2Ms with MLPs and HGNNs across six heterogeneous graph datasets under the standard transductive setting. As shown in Table \ref{Tab: Transductive Setting}, both HG2M and HG2M+ outperform vanilla MLPs by significant margins. Compared to HGNNs, HG2M shows a slight performance degradation, while HG2M+ achieves the best performance on 5/6 datasets. Compared to HG2M, HG2M+ improves the performance by 2.36\% on average across different datasets, which demonstrates the effectiveness of distilling reliable and heterogeneous semantic knowledge.

To further assess the performance of HG2Ms, we conduct experiments in a realistic production (\textit{prod}) scenario that contains both inductive (\textit{ind}) and transductive (\textit{tran}) settings, as detailed in Section \ref{Sec: Transductive vs. Inductive}. We report \textit{tran}, \textit{ind} results, and interpolated \textit{prod} results in Table \ref{Tab: Production Setting}. The prod results provide a clearer insight into the model’s generalization ability and accuracy in production environments. As we can see, HG2M and HG2M+ show average accuracy improvements of 6.24\% and 7.78\%, respectively, over vanilla MLPs. Additionally, HG2M+ outperforms HG2M by an average of 1.54\% in the \textit{prod} setting across all datasets. The \textit{prod} results of HG2Ms are also competitive with those of teacher HGNNs, suggesting that HG2Ms can be deployed as a much faster model with no or only slight performance degradation.

On the TMDB, ArXiv, and IGB-3M-19 datasets, HG2M+ performs slightly worse than HGNNss. We hypothesize that this is due to a significant distribution shift between the training and test nodes on these datasets, which makes it challenging for HG2Ms to capture the underlying patterns without leveraging neighbor information, as HGNNs do. Nonetheless, it is worth noting that HG2Ms consistently outperform vanilla MLPs across all datasets.

\subsection{How do HG2Ms compare to other GNNs-to-MLPs?(RQ2)}\label{Sec: GNN2MLP Comparison}
While previous GNN-to-MLP methods have mainly focused on homogeneous graphs, applying them to heterogeneous graphs may lead to suboptimal performance, as they fail to consider the intrinsic heterogeneity and semantic richness captured by meta-path structures. Thus, we meticulously design HG2M+, which integrates RND and RMPD, to distill reliable and heterogeneous knowledge from HGNNs to MLPs.


\begin{table}[!ht]
    \begin{center}
    \setlength{\tabcolsep}{4pt}
    {\caption{Transductive accuracy of different distillation methods.}\label{Tab: GNN2MLP Comparison}}
    \begin{tabular}{c|ccc}
    \toprule
    Method & TMDB & CroVal & IGB-549K-19 \\ \midrule
    HG2M/GLNN & 80.64±0.44 & 86.16±0.20 & 57.99±0.49 \\
    RKD-MLP & 83.92±0.45 & 87.81±0.32 & 58.90±0.64 \\
    FF-G2M & 83.80±0.50 & 87.37±0.55 & 58.42±0.63 \\
    KRD & 84.01±0.30 & 88.37±0.47 & 58.84±0.72 \\
    NOSMOG & 84.09±1.53 & \underline{88.69±1.31} & \underline{59.43±1.20} \\
    HIRE & 82.80±0.99 & 88.01±0.82 & 58.72±1.01 \\
    TGS & \underline{84.21±0.41} & 88.44±0.58 & 59.02±0.81 \\
    TeKAP & 80.99±0.38 & 87.39±0.61 & 58.22±0.77 \\
    HG2M+ & \textbf{84.69±0.39} & \textbf{89.05±0.10} & \textbf{59.62±0.49} \\
    \bottomrule
    \end{tabular}
    \end{center}
\end{table}

\begin{table*}[!ht]
    \begin{center}
    {\caption{Inductive inference time (in ms) on 5 randomly chosen nodes. NS-$n$ means inference neighbor sampling with fan-out $n$ for each edge type.}\label{Tab: Inference Time}}
    \begin{tabular}{c|c|cccc|c}
    \toprule
    Dataset & RSAGE & NS-20 & NS-15 & NS-10 & NS-5 & HG2Ms \\ \midrule
    TMDB & 10.75 & 12.00 (0.90×) & 11.83 (0.91×) & 10.94 (0.98×) & 10.45 (1.03×) & \textbf{0.27 (39.81×)} \\
    CroVal & 14.17 & 13.08 (1.08×) & 12.59 (1.13×) & 11.26 (1.26×) & 10.93 (1.30×) & \textbf{0.25 (56.68×)} \\
    ArXiv & 26.21 & 9.15 (2.86×) & 8.94 (2.93×) & 7.36 (3.56×) & 7.17 (3.66×) & \textbf{0.23 (113.96×)} \\
    IGB-549K-19 & 16.49 & 16.32 (1.01×) & 15.89 (1.04×) & 14.63 (1.13×) & 14.00 (1.18×) & \textbf{0.26 (63.42×)} \\
    IGB-549K-2K & 16.87 & 16.13 (1.05×) & 16.05 (1.05×) & 14.88 (1.13×) & 14.11 (1.20×) & \textbf{0.34 (49.62×)} \\
    IGB-3M-19 & 125.15 & 19.38 (6.46×) & 18.62 (6.72×) & 16.22 (7.72×) & 15.07 (8.30×) & \textbf{0.33 (379.24×)} \\
    \bottomrule
    \end{tabular}
    \end{center}
\end{table*}

To validate the effectiveness of HG2M+, we further compare it with seven previous knowledge distillation frameworks: RKD-MLP \cite{RKD-MLP}, FF-G2M \cite{FF-G2M}, KRD \cite{KRD}, NOSMOG \cite{NOSMOG}, HIRE \cite{HIRE}, TGS \cite{TGS}, and TeKAP \cite{TeKAP}, which leverage more complex techniques like ensemble distillation, feature-based distillation, and representational similarity distillation. Table \ref{Tab: GNN2MLP Comparison} presents the performance of these models on the TMDB, CroVal, and IGB-549K-19 datasets. We can find that HG2M+ consistently achieves the best performance. This highlights the superiority of our HG2M+ for heterogeneous graphs.

\subsection{How efficient are HG2Ms compared to HGNNs? (RQ3)}
Inference efficiency and accuracy stand as two pivotal criteria for evaluating a machine learning system. Nowadays, growing demands of industrial graph learning applications necessitate models capable of low-latency inference. Here we conduct a comparison of inference times among RSAGE, RSAGE with neighbor sampling (NS), and HG2Ms on 5 randomly selected nodes. NS-$n$ represents that each node takes messages from $n$ sampled neighbors per edge type during inference. As shown in Table \ref{Tab: Inference Time}, our HG2Ms are considerably faster than the baseline methods, achieving speedups ranging from 39.81× to 379.24× compared to the teacher RSAGE. This improvement can be attributed to the fact that HG2Ms, which are essentially well-trained MLPs, avoid the computationally intensive multiplication-and-accumulation operations over the features of numerous neighbors in HGNNs.  These results demonstrate the superior inference efficiency of our HG2Ms, making them particularly well-suited for latency-sensitive applications.

We further present an extended efficiency analysis on the largest dataset IGB-3M-19, and report the time cost and GPU memory usage (in MB) for both training and inference phases. The inference results are also based on 5 randomly selected test nodes. As shown in Table \ref{Tab: Efficiency Analysis}, despite incurring slightly higher training time than RSAGE (within 5\%), HG2Ms require less than half the training memory. Most notably, at inference time, HG2Ms achieve 379× speedup and 28× lower memory usage while maintaining competitive or better accuracy than RSAGE. These results further highlight the practical advantages of our framework for deployment in large-scale and memory-constrained settings.

\begin{table}[!ht]
    \begin{center}
    \setlength{\tabcolsep}{2pt}
    {\caption{Comprehensive efficiency analysis on IGB-3M-19.}\label{Tab: Efficiency Analysis}}
    \begin{tabular}{c|cc|cc}
    \toprule
    Method & Train Time (s) & Train Mem. & Infer. Time (ms) & Infer. Mem. \\ \midrule
    MLP & 455.62 & 5401.76 & 0.33 & 9.66 \\
        RSAGE & 3500.21 & 23784.22 & 125.15 & 275.70 \\
        HG2M & 3615.88 (0.97×) & 9650.93 (2.46×) & 0.33 (379.24×) & 9.66 (28.54×) \\
        HG2M+ & 3698.23 (0.95×) & 11591.48 (2.05×) & 0.33 (379.24×) & 9.66 (28.54×) \\
    \bottomrule
    \end{tabular}
    \end{center}
\end{table}


\subsection{How does each component contribute to HG2M+? (RQ4)}\label{Sec: Ablation Study}
Since HG2M+ contains two essential components (i.e., reliable node distillation (RND) and reliable meta-path distillation (RMPD)), we conduct ablation studies to assess the individual contributions of each component when integrated independently into HG2M. Table \ref{Tab: Ablation Study} demonstrates that performance improves with the addition of each component, indicating their effectiveness. In general, the incorporation of RND contributes a lot, as RND can filter out noisy information, making the heterogeneous semantic knowledge distilled by RMPD more reliable. 

\begin{table}[!ht]
    \begin{center}
    {\caption{Ablation study for independent components in HG2M+. The increasing performance shows that each component in HG2M+ contributes positively to the KD process.}\label{Tab: Ablation Study}}
    \begin{tabular}{c|cccc}
    \toprule
    Dataset & HG2M & w/ RND & w/ RMPD & HG2M+ \\ \midrule
    TMDB & 80.64±0.44 & 83.67±0.38 & 83.29±0.33 & \textbf{84.69±0.39} \\
    CroVal & 86.16±0.20 & 88.05±0.12 & 88.37±0.06 & \textbf{89.05±0.10} \\
    ArXiv & 76.46±0.10 & 78.15±0.32 & 77.68±0.28 & \textbf{78.80±0.30} \\
    IGB-549K-19 & 57.99±0.49 & 59.24±0.47 & 58.72±0.50 & \textbf{59.62±0.49} \\
    IGB-549K-2K & 57.34±0.50 & 58.25±0.45 & 57.71±0.60 & \textbf{59.20±0.66} \\
    IGB-3M-19 & 64.89±0.40 & 65.47±0.37 & 65.12±0.23 & \textbf{66.27±0.40} \\
    \bottomrule
    \end{tabular}
    \end{center}
\end{table}

\begin{figure}[ht]
    \centering
    \subfigure[Accuracy of Selected Reliable Nodes]{\includegraphics[width=0.49\linewidth]{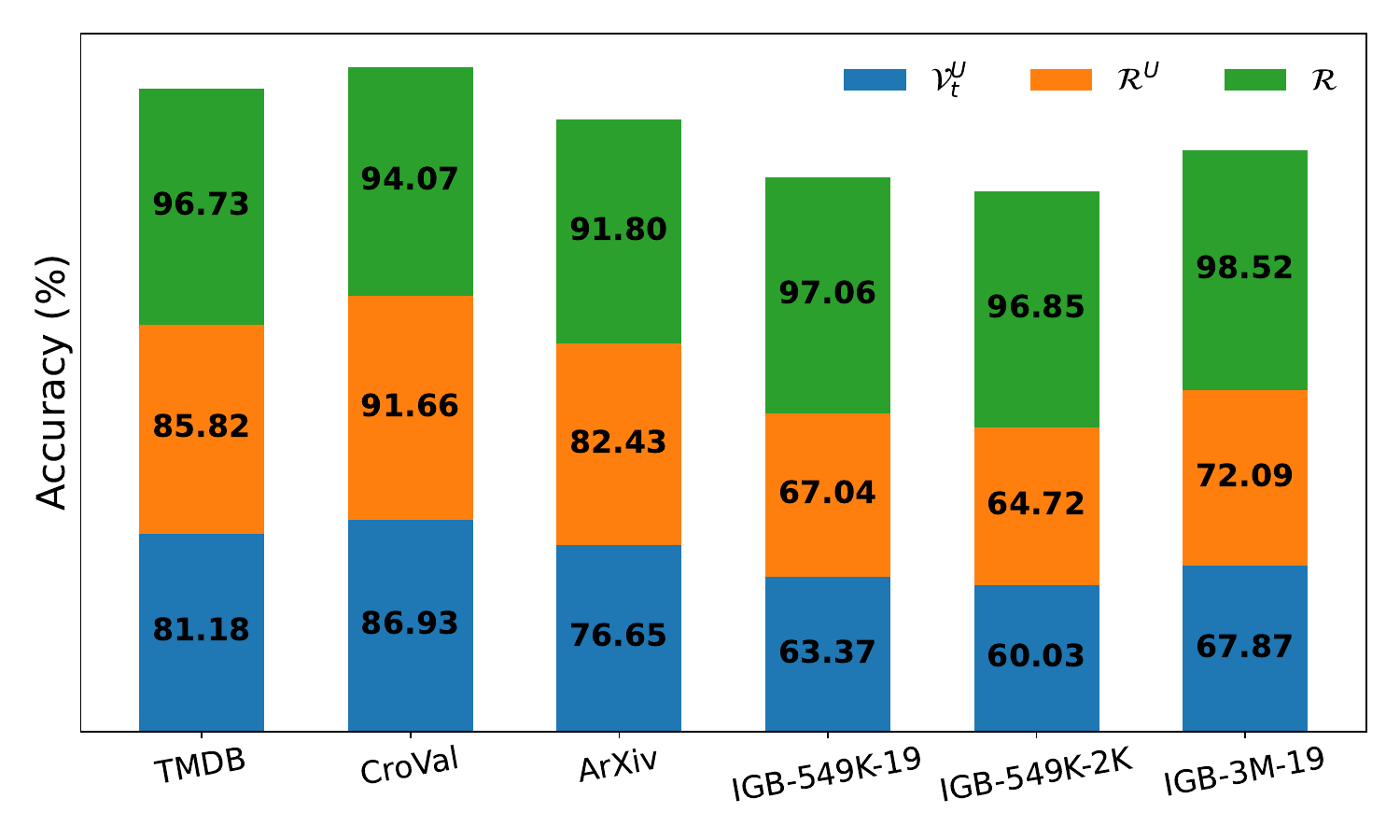}}
    \subfigure[Homophily of Selected Meta-path-based Neighbor Pairs]{\includegraphics[width=0.49\linewidth]{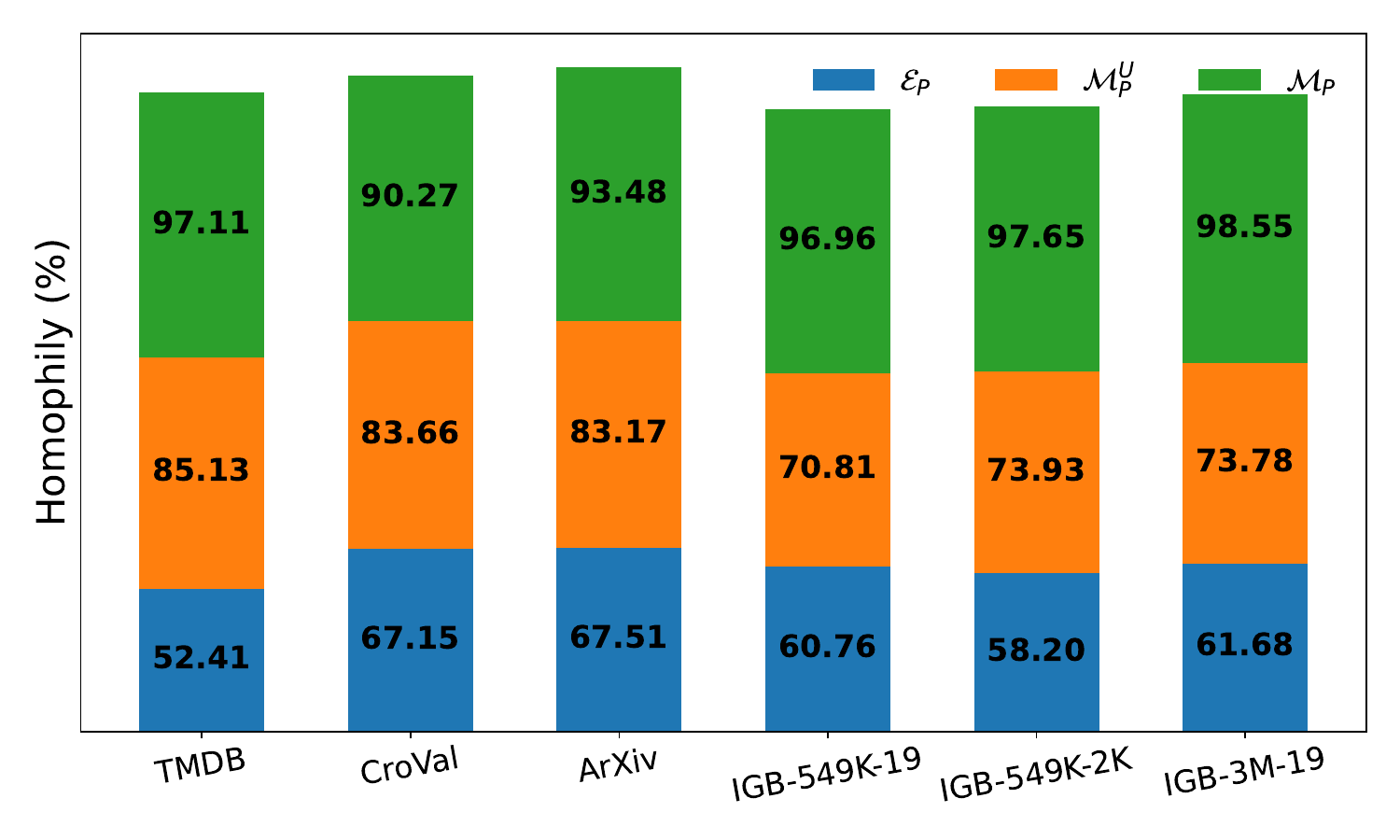}}
    \caption{Effectiveness of our reliable node selection in RND and intra-class meta-path-based neighbor estimation in RMPD. 
    (a) The accuracy of selected reliable nodes $\mathcal{R}^U$ (orange) is much higher than that of the raw unlabeled node set $\mathcal{V}_t^U$ (blue). (b) The homophily (i.e., the fraction of intra-class pairs) of selected meta-path-based neighbor pairs $\mathcal{M}_P^U$ (orange) is much higher than that of the raw meta-path-based neighbor pair set $\mathcal{E}_P$ (blue).
    }\label{Fig: Ablation RND RMPD}
\end{figure}

\begin{figure*}[ht]
    \centering
    \subfigure[TMDB]{\includegraphics[width=0.325\linewidth]{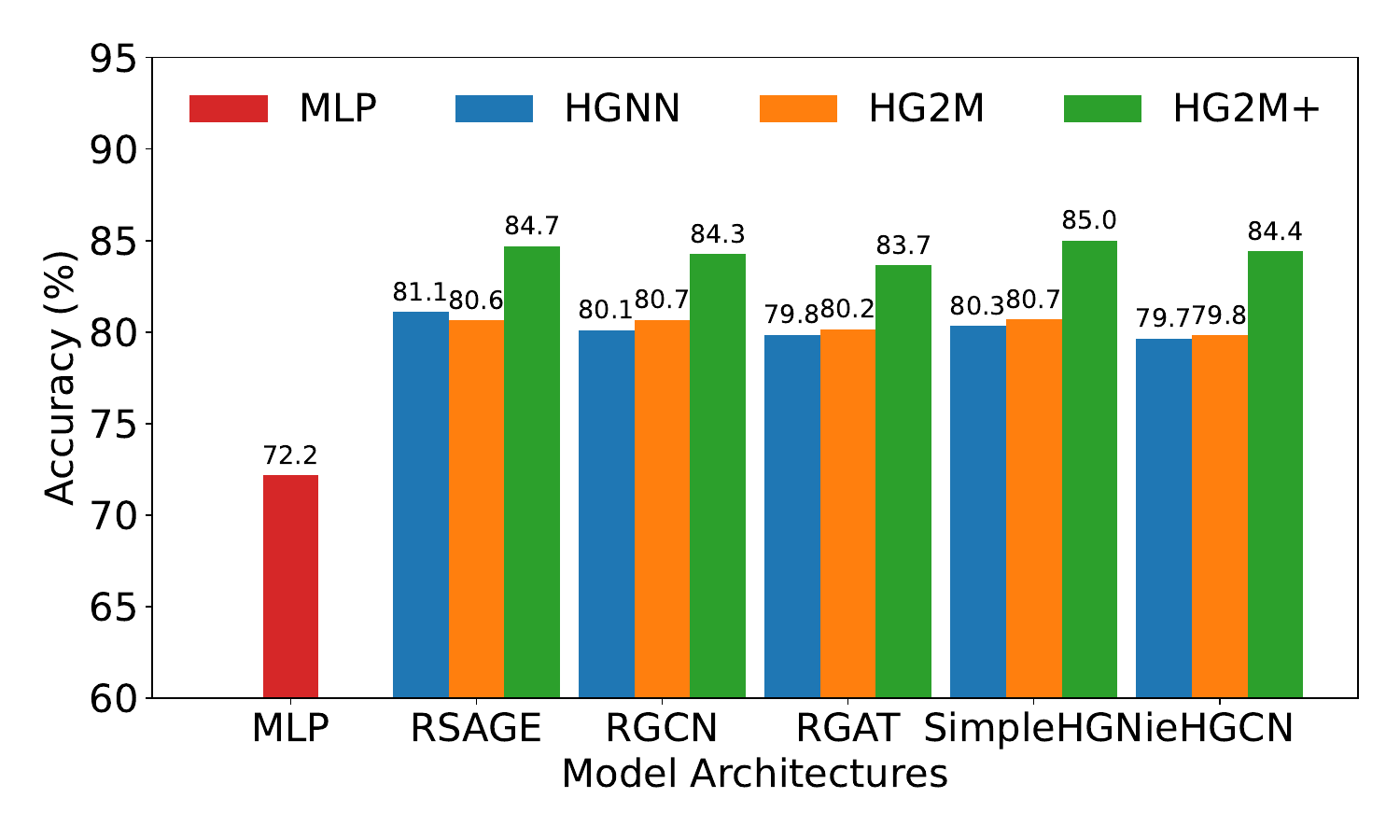}}
    \subfigure[CroVal]{\includegraphics[width=0.325\linewidth]{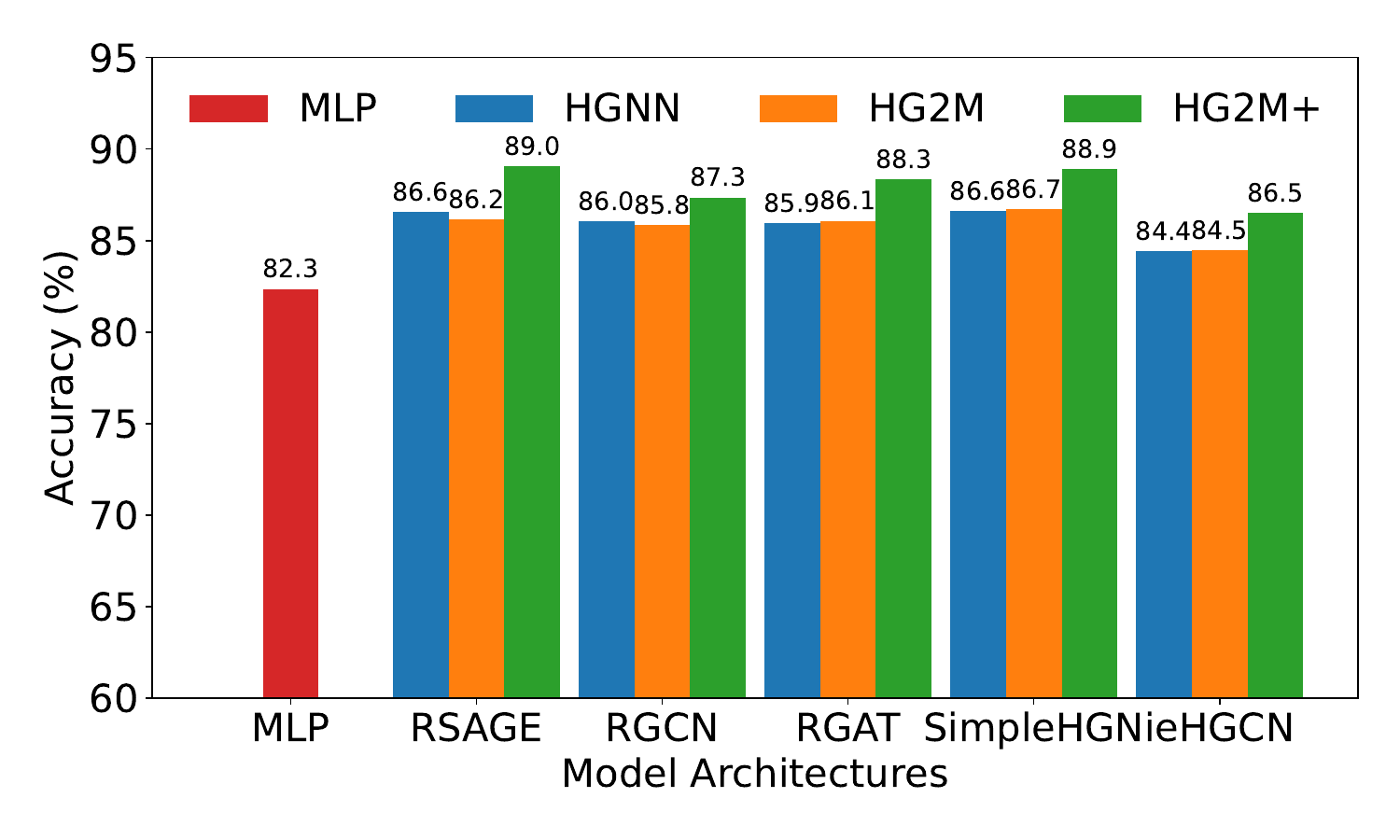}}
    \subfigure[IGB-549K-19]{\includegraphics[width=0.325\linewidth]{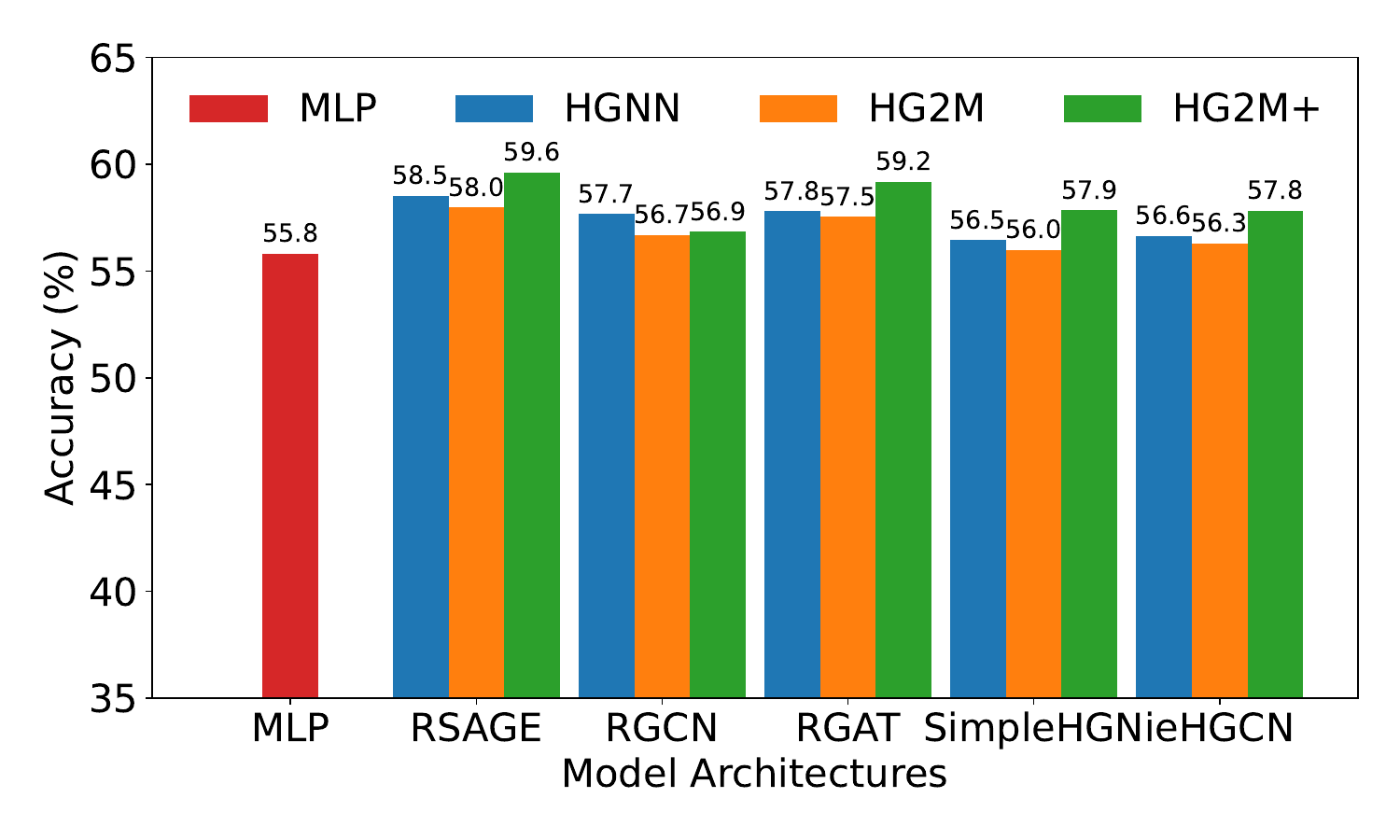}}
    \caption{Transductive Accuracy vs. Teacher HGNN Architectures. HG2Ms can learn from different HGNN teachers to improve over MLPs and achieve comparable results to teachers.}\label{Fig: Param Model Architecture}
\end{figure*}

\begin{figure*}[ht]
    \centering
    \subfigure[TMDB]{\includegraphics[width=0.325\linewidth]{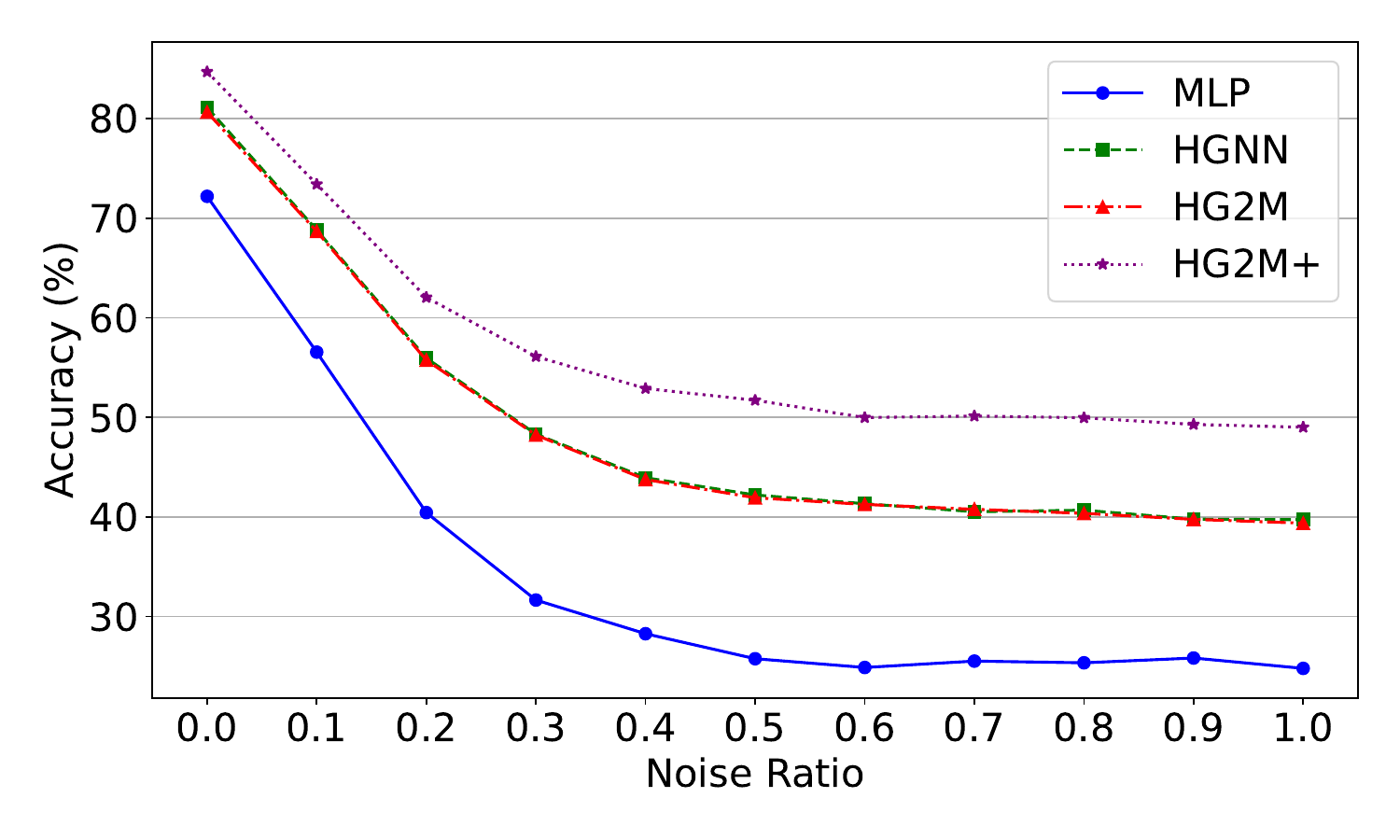}}
    \subfigure[CroVal]{\includegraphics[width=0.325\linewidth]{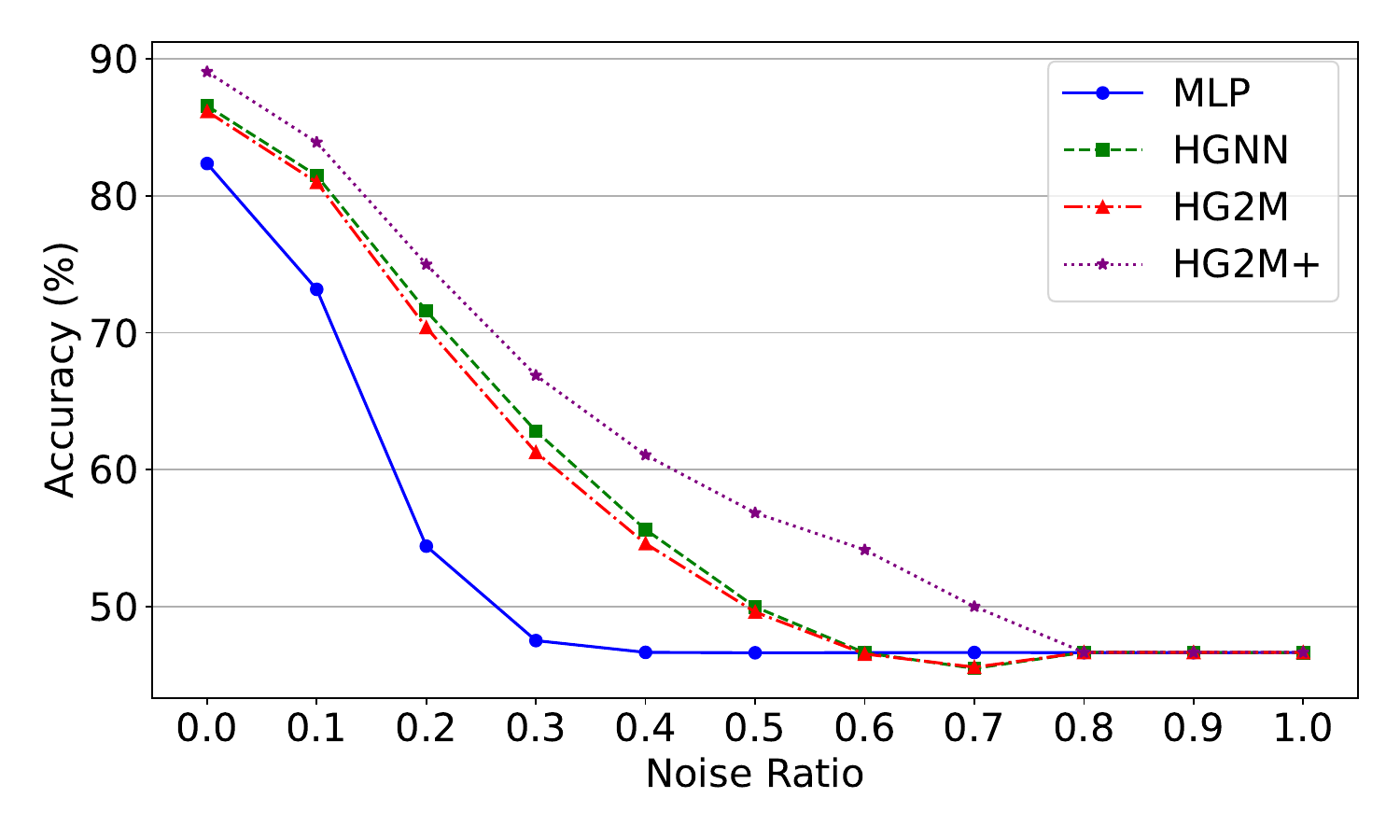}}
    \subfigure[IGB-549K-19]{\includegraphics[width=0.325\linewidth]{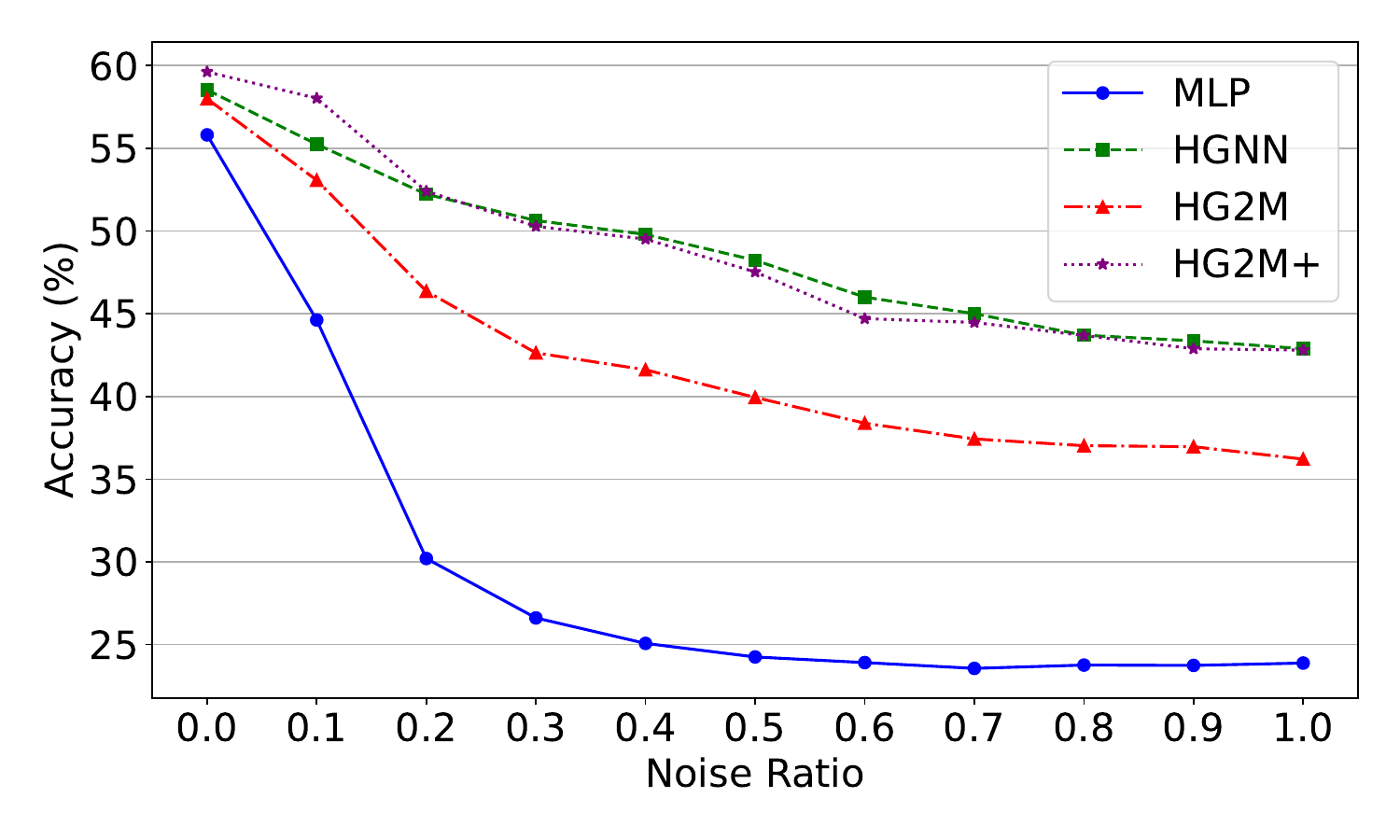}}
    \caption{Transductive Accuracy vs. Node Feature Noise. HG2Ms have comparable performance to HGNNs. Adding more noise decreases HG2Ms' performance slower than MLPs.} \label{Fig: Param Noise Ratio}
\end{figure*}

\begin{figure*}[ht]
    \centering
    \subfigure[TMDB]{\includegraphics[width=0.325\linewidth]{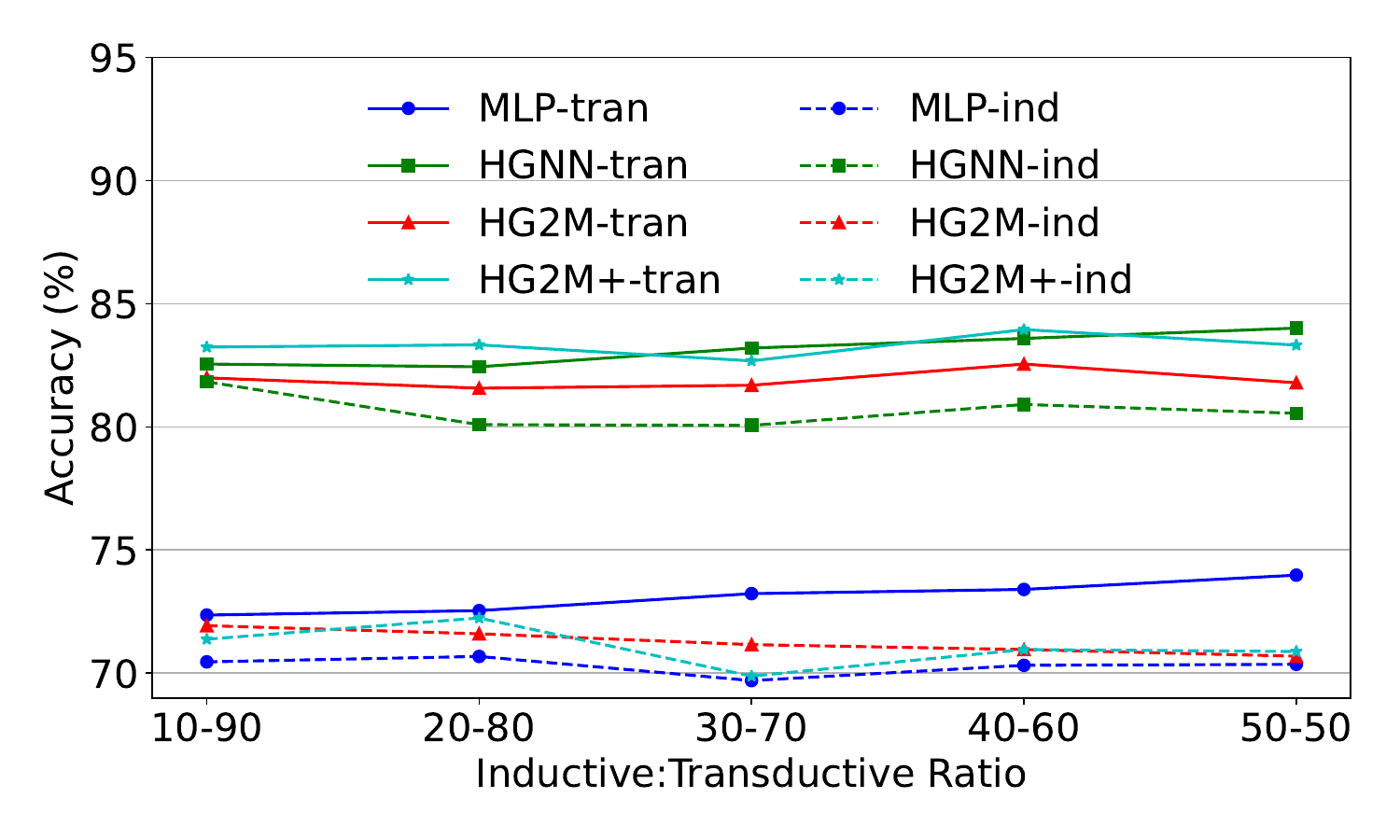}}
    \subfigure[CroVal]{\includegraphics[width=0.325\linewidth]{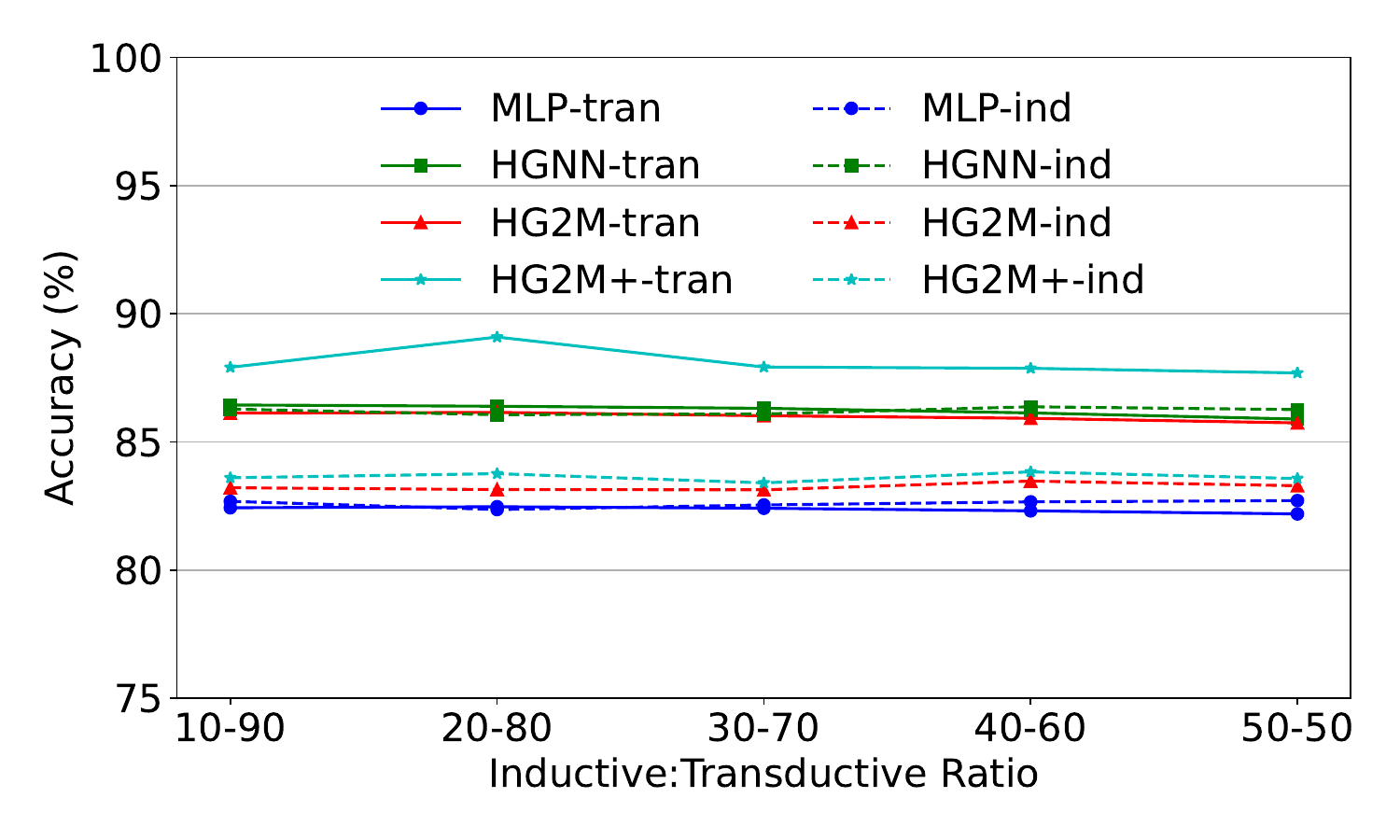}}
    \subfigure[IGB-549K-19]{\includegraphics[width=0.325\linewidth]{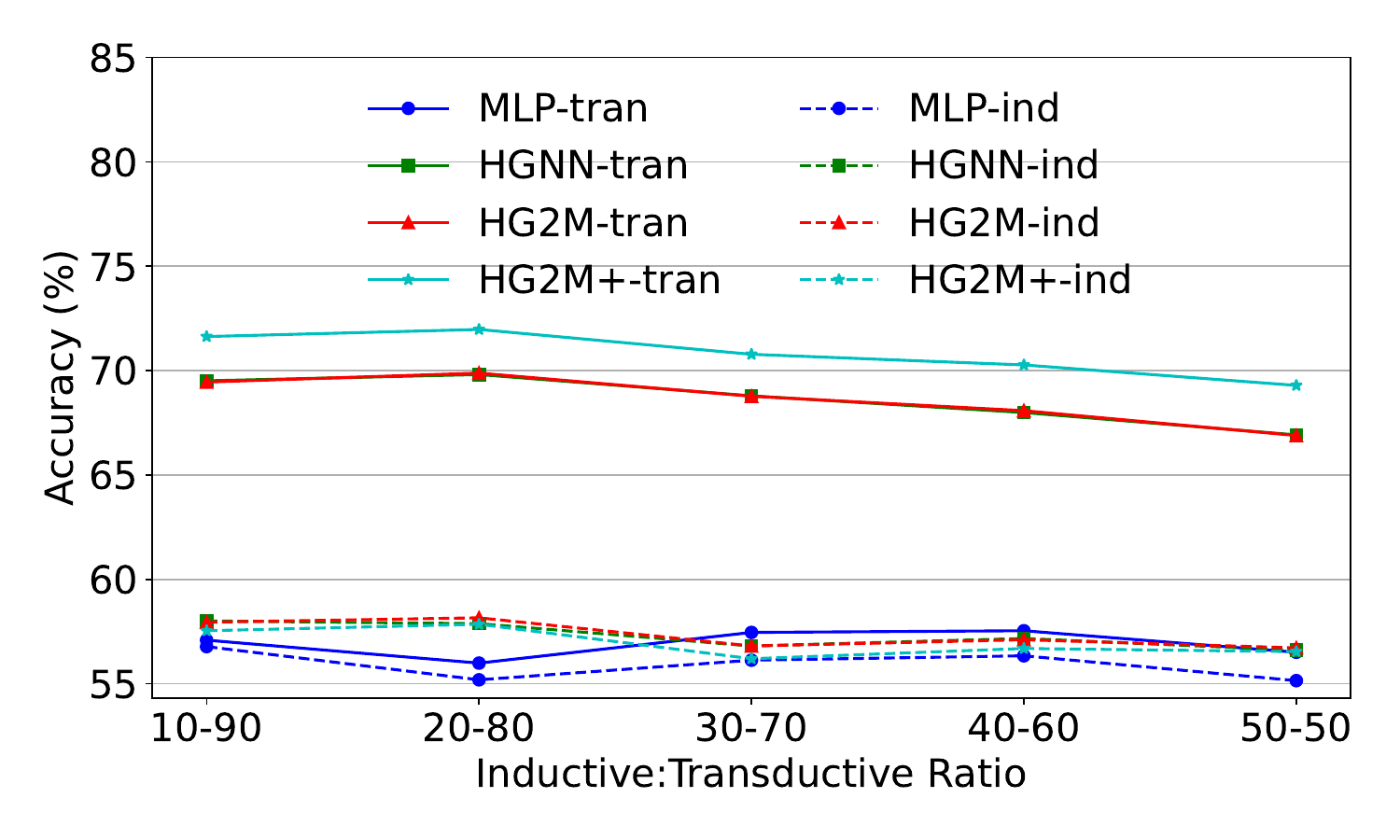}}
    \caption{Accuracy vs. Inductive:Transductive Ratio under the production setting. Altering the inductive split rate doesn't affect the accuracy much.} \label{Fig: Param Split Rate}
\end{figure*}

We also conduct an in-depth analysis to demonstrate the effectiveness of our reliable node selection strategy in RND and our intra-class meta-path-based neighbor estimation strategy in RMPD. Figure \ref{Fig: Ablation RND RMPD}(a) presents the accuracy of raw unlabelled nodes $\mathcal{V}_t^U$, selected unlabelled reliable nodes $\mathcal{R}^U$, and all selected reliable nodes $\mathcal{R}$ with a selection proportion of $p=0.9$. Figure \ref{Fig: Ablation RND RMPD}(b) displays the homophily of all meta-path-$P$-based neighbor pairs $\mathcal{E}_P$, selected unlabelled meta-path-$P$-based neighbor pairs $\mathcal{M}_P^U$, and all selected meta-path-$P$-based neighbor pairs $\mathcal{M}_P$. Here, homophily represents the fraction of intra-class neighbor pairs, with meta-paths MAM, QTQ, PAP, and PAP applied to TMDB, CroVal, ArXiv, and IGBs, respectively. We observed that the accuracy of selected unlabelled reliable nodes (orange) is significantly higher than that of raw unlabelled reliable nodes (blue), and the homophily of selected unlabelled meta-path-$P$-based neighbor pairs (orange) is also markedly higher than that of all meta-path-$P$-based neighbor pairs (blue). Furthermore, the accuracy of the final reliable nodes (green) used for distillation and the homophily of the final reliable intra-class meta-path-based neighbor pairs (green) used for distillation both exceed 0.9. These findings underscore the efficacy of our reliable node selection strategy in RND for identifying reliable nodes and our intra-class meta-path-based neighbor estimation strategy in RMPD for identifying intra-class neighbors.

\begin{table}[!ht]
    \begin{center}
    {\caption{Ablation study on the effect of different meta-paths in RMPD.}\label{Tab: MetaPath}}
    \begin{tabular}{c|cccc}
    \toprule
    Dataset & w/ None & w/ MP1 & w/ MP2 & w/ Both \\ \midrule
    TMDB & 83.67±0.38 & 84.38±0.32 & 84.29±0.33 & \textbf{84.69±0.39} \\
    CroVal & 88.05±0.12 & 88.37±0.06 & 88.76±0.04 & \textbf{89.05±0.10} \\
    IGB-549K-19 & 59.24±0.47 & 59.56±0.58 & 59.42±0.50 & \textbf{59.62±0.49} \\
    \bottomrule
    \end{tabular}
    \end{center}
\end{table}

\subsection{How does each meta-path contribute to HG2M+? (RQ5)}
In our main experiments, we use all meta-paths listed in Table \ref{Tab: Dataset Statistics} for RMPD. To better understand the contribution of each meta-path and examine possible redundancy, we conduct ablation studies on three representative datasets: TMDB, CroVal, and IGB-549K-19. Specifically, we compare four settings: (1) without using any meta-path (w/ None), (2) using only the first meta-path in Table \ref{Tab: Dataset Statistics} (w/ MP1, i.e., HG2M+ w/o RMPD), (3) using only the second meta-path (w/ MP2), and (4) using both (w/ Both, i.e., HG2M+). The results shown in Table \ref{Tab: MetaPath} demonstrate that each individual meta-path improves performance over the baseline without RMPD, and using both meta-paths yields the best results, confirming that each meta-path contributes unique, non-redundant information. Thus, meta-paths are complementary and beneficial to the overall model performance.

\subsection{How do HG2Ms perform with different teachers? (RQ6)}\label{Sec: Model Architecture}
We have adopted RSAGE to represent the teacher HGNNs so far. However, since different HGNN architectures may exhibit varying performances across datasets, we investigate whether HG2Ms can perform well when trained with other teacher HGNNs. In Figure \ref{Fig: Param Model Architecture}, we present the transductive performance of HG2Ms when distilled from different teacher HGNNs, including RGCN, RGAT, SimpleHGN, and ieHGCN, on TMDB, CroVal, and IGB-549K-19 datasets. We see that HG2Ms can effectively learn from different teachers and outperform vanilla MLPs. HG2M achieves comparable performance to teachers, while HG2M+ consistently surpasses them, underscoring the efficacy of our proposed model. 


\begin{figure*}[ht]
    \centering
    \subfigure[TMDB]{\includegraphics[width=0.32\linewidth]{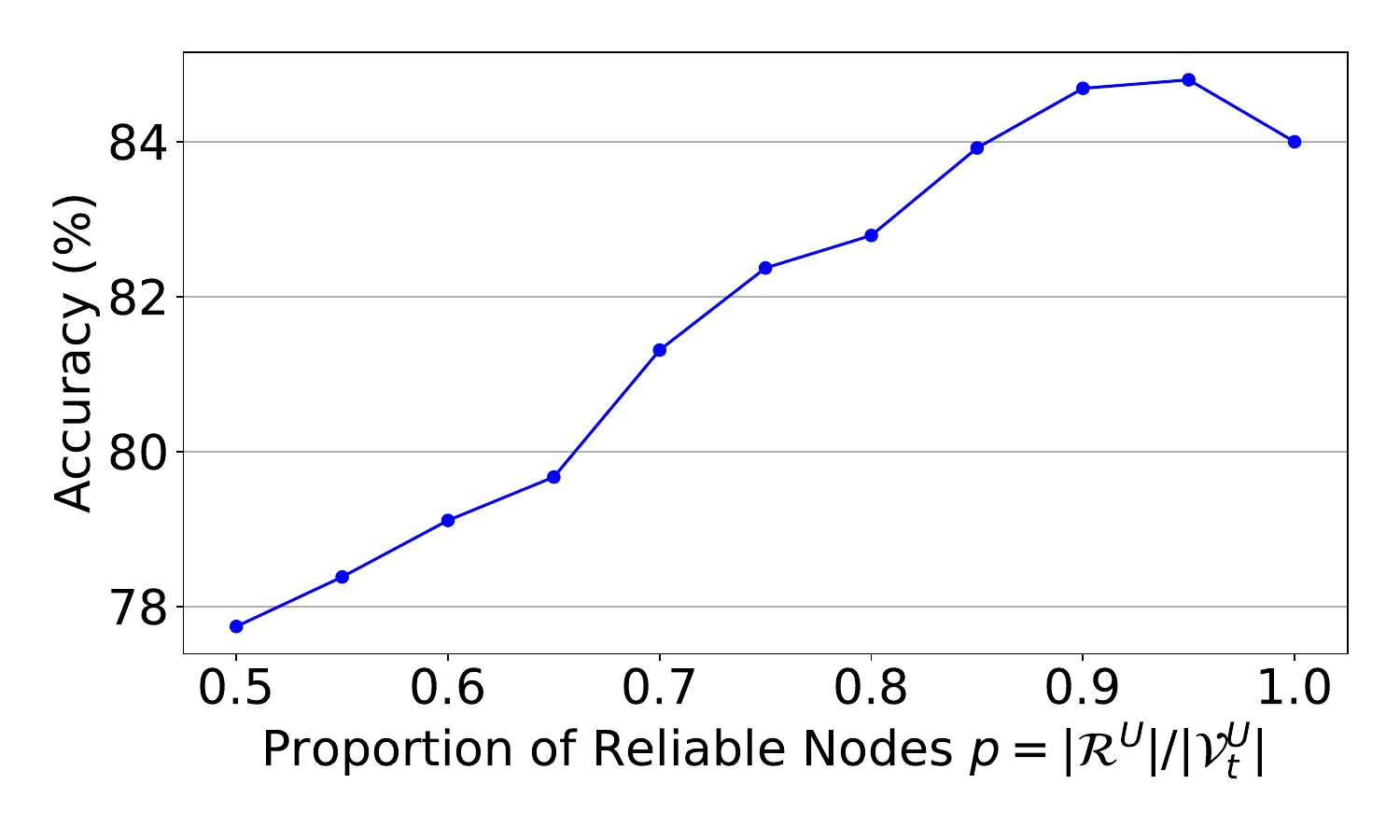}}
    \subfigure[CroVal]{\includegraphics[width=0.32\linewidth]{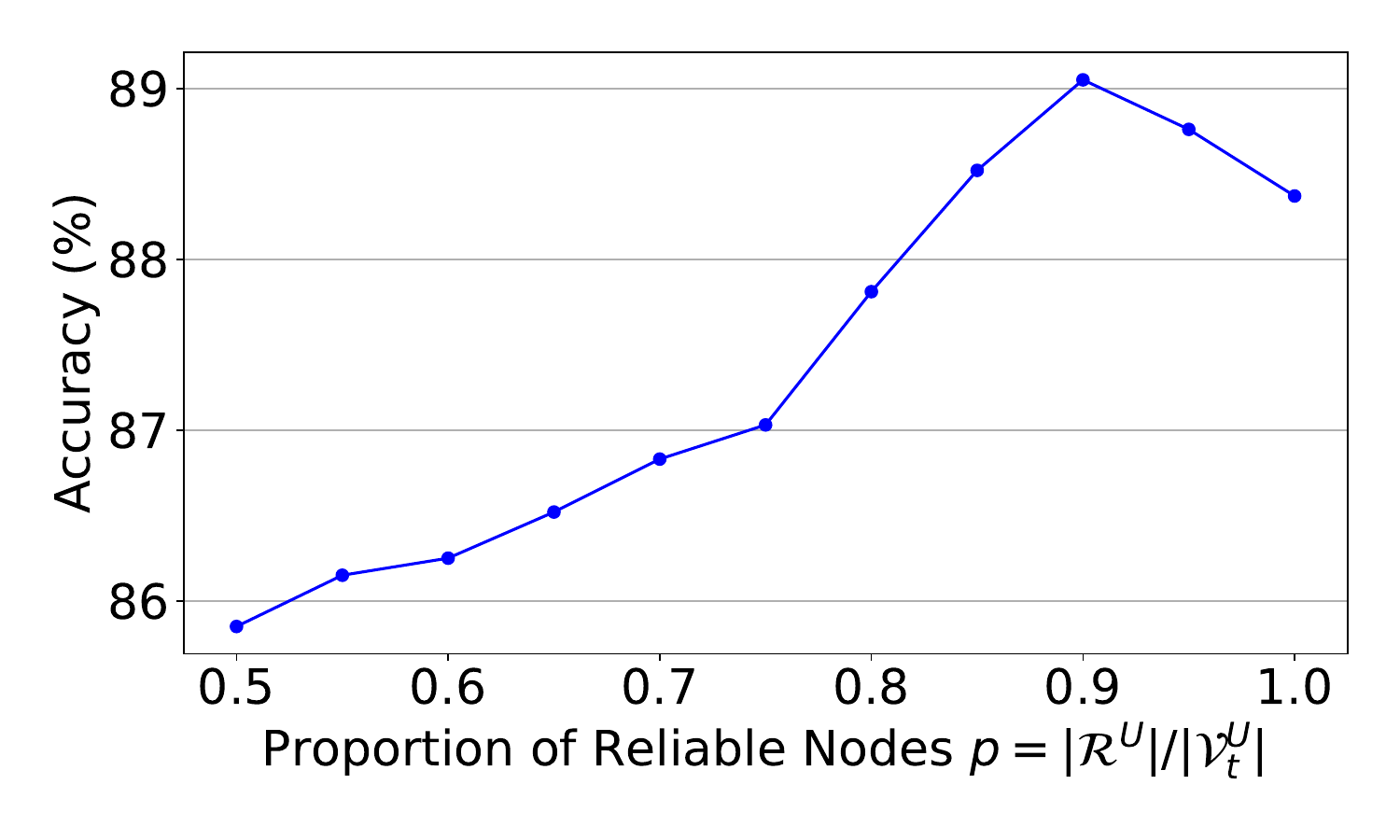}}
    \subfigure[IGB-549K-19]{\includegraphics[width=0.32\linewidth]{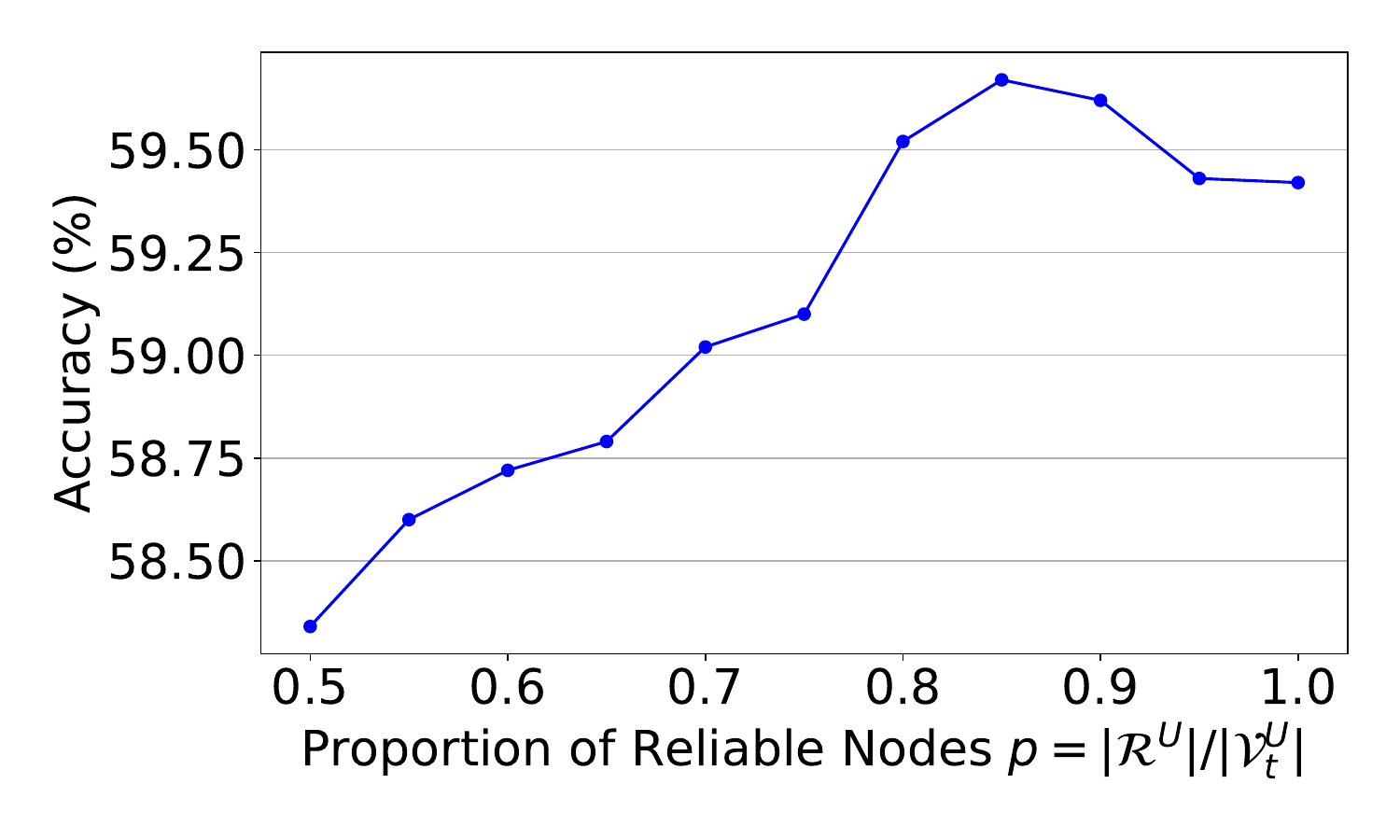}}
    \caption{Impact of the proportion of selected reliable nodes on the performance of HG2M+.}\label{Fig: Param Proportion P1}
\end{figure*}

\begin{figure*}[ht]
    \centering
    \subfigure[TMDB]{\includegraphics[width=0.32\linewidth]{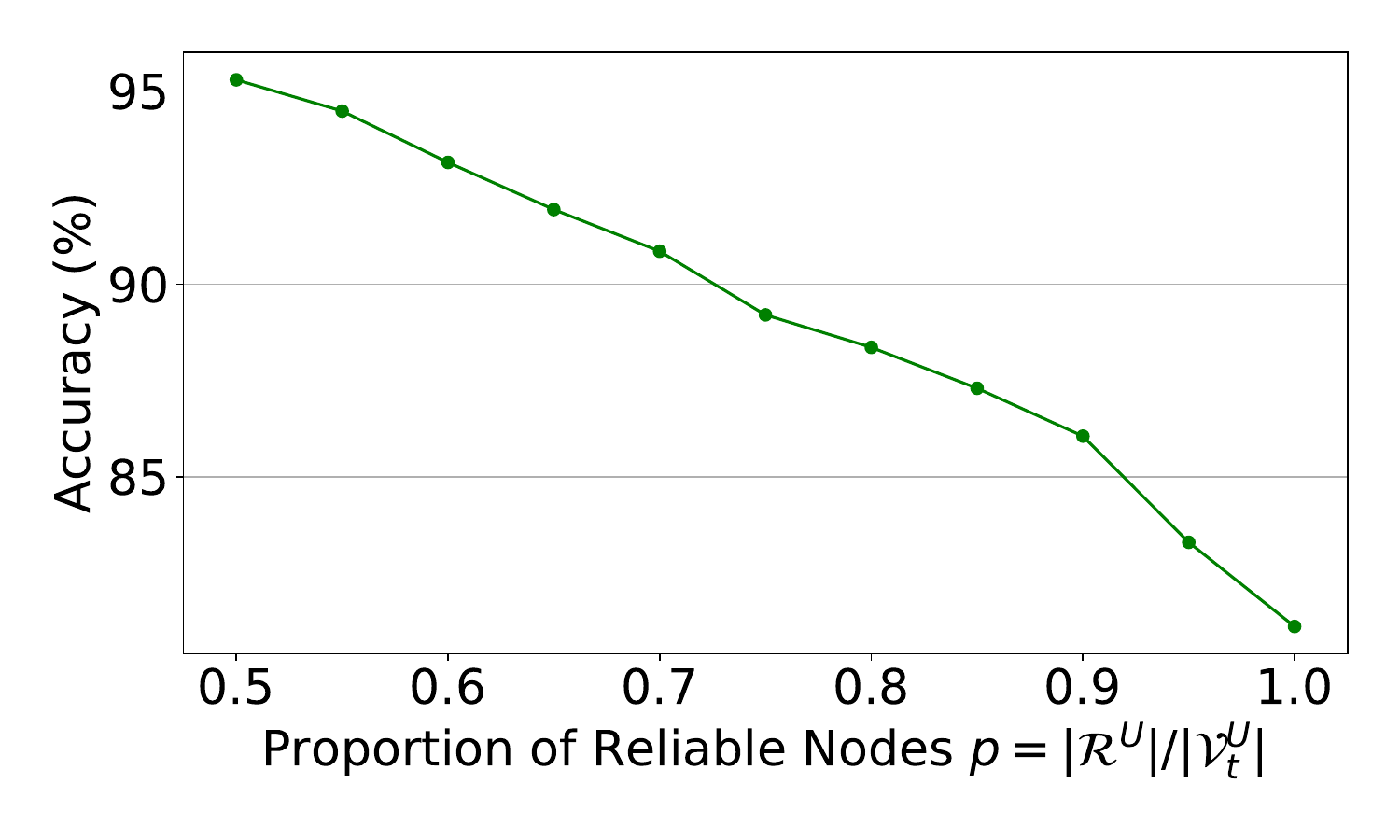}}
    \subfigure[CroVal]{\includegraphics[width=0.32\linewidth]{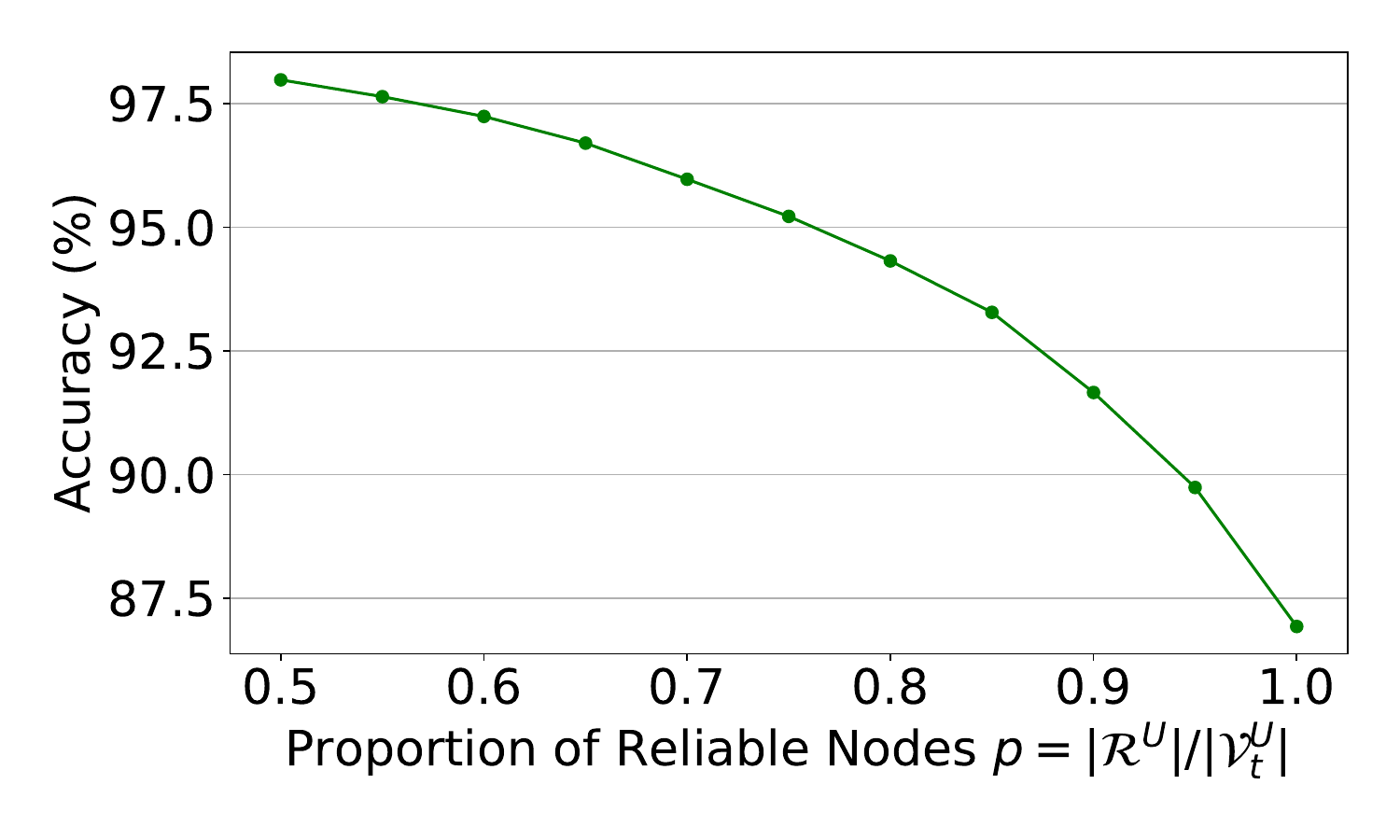}}
    \subfigure[IGB-549K-19]{\includegraphics[width=0.32\linewidth]{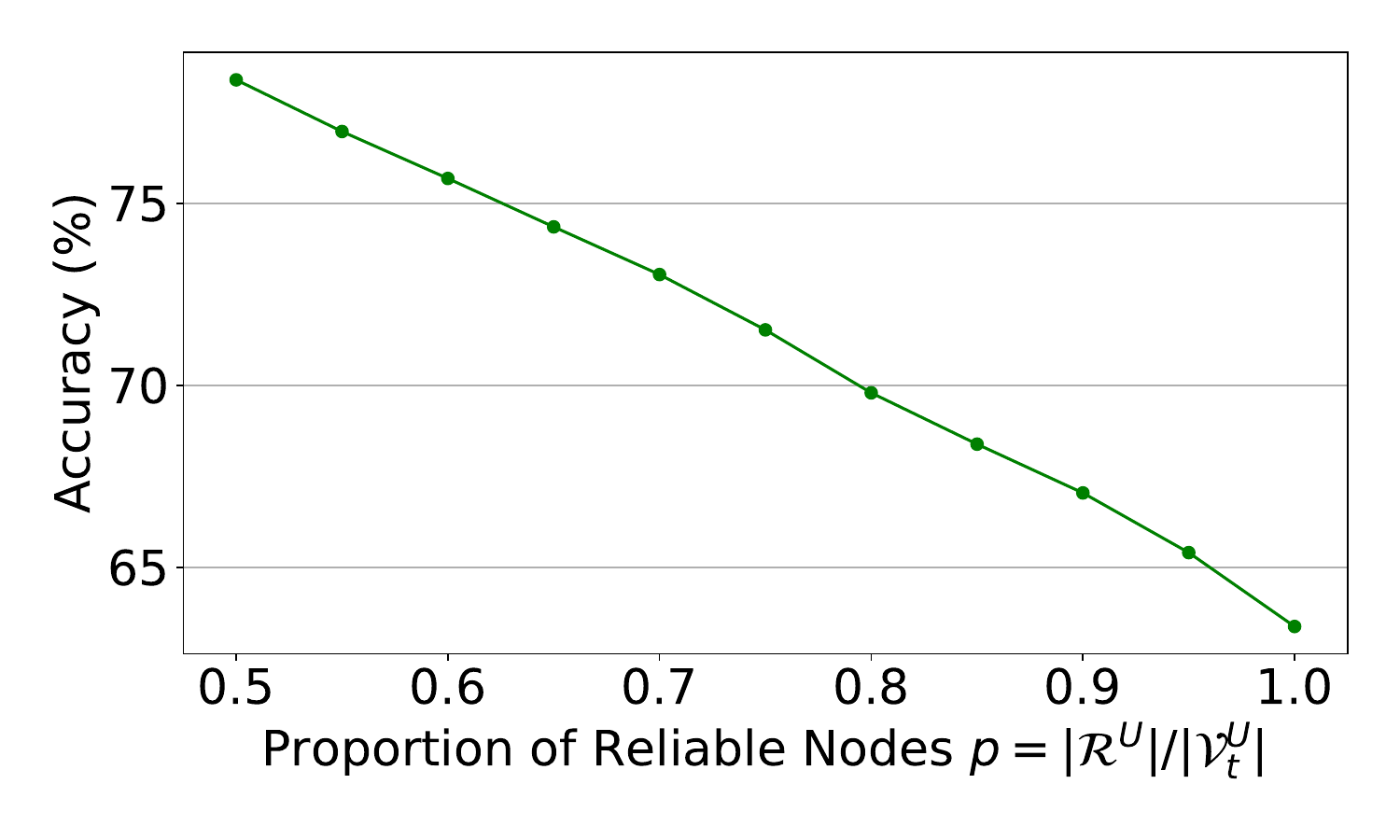}}
    \caption{Impact of the proportion of selected reliable nodes on the accuracy of selected reliable nodes.}\label{Fig: Param Proportion P2}
\end{figure*}

\begin{figure*}[ht]
    \centering
    \subfigure[TMDB]{\includegraphics[width=0.32\linewidth]{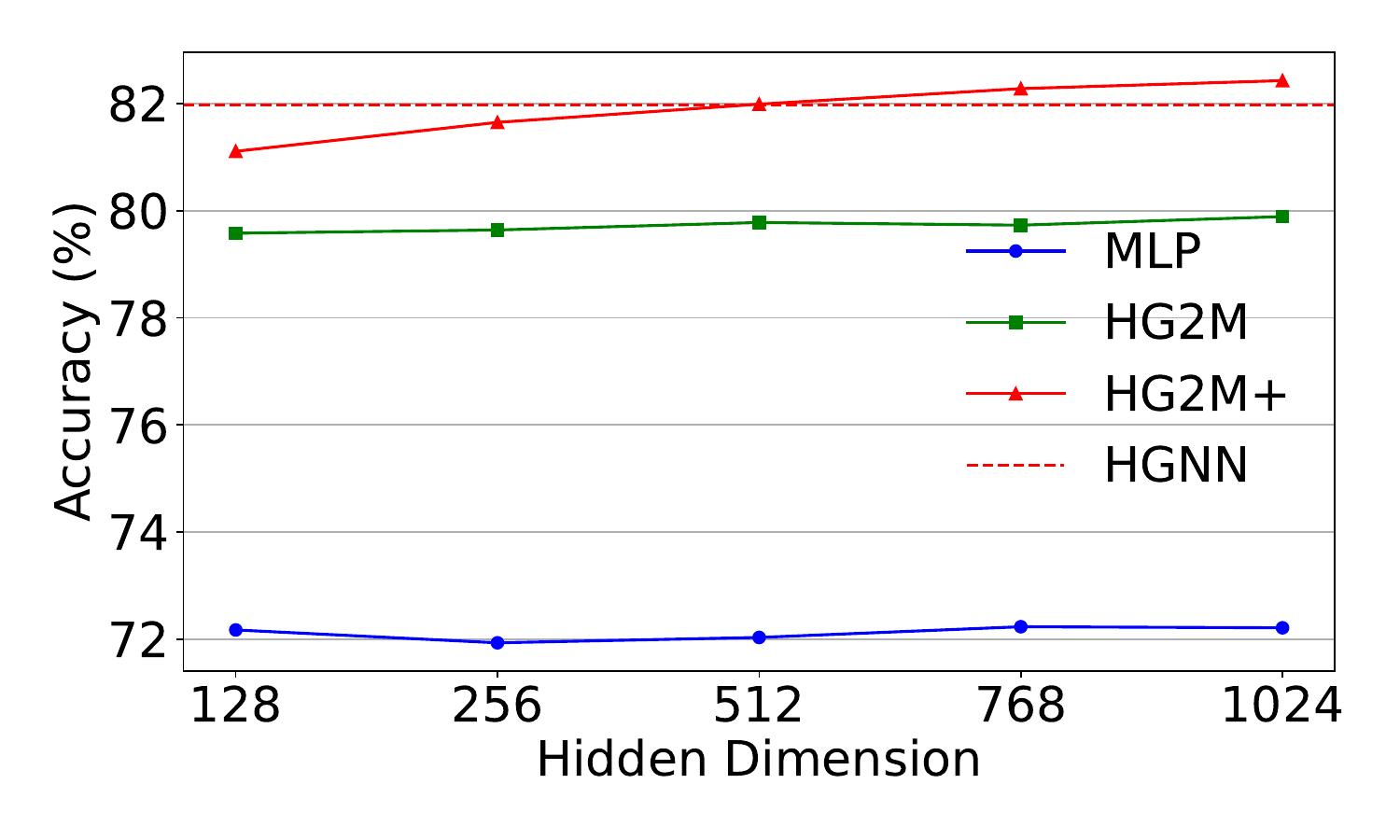}}
    \subfigure[ArXiv]{\includegraphics[width=0.32\linewidth]{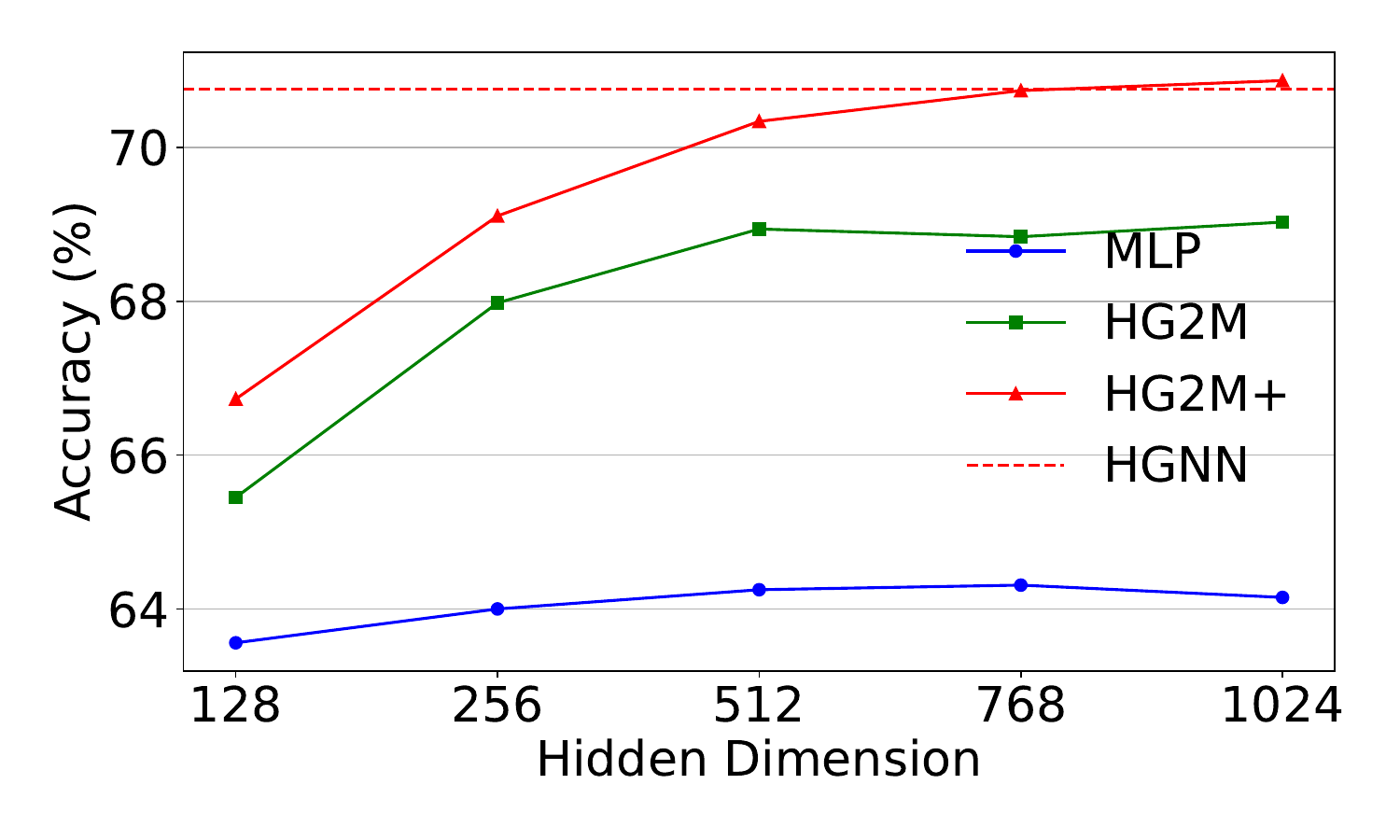}}
    \subfigure[IGB-549K-19]{\includegraphics[width=0.32\linewidth]{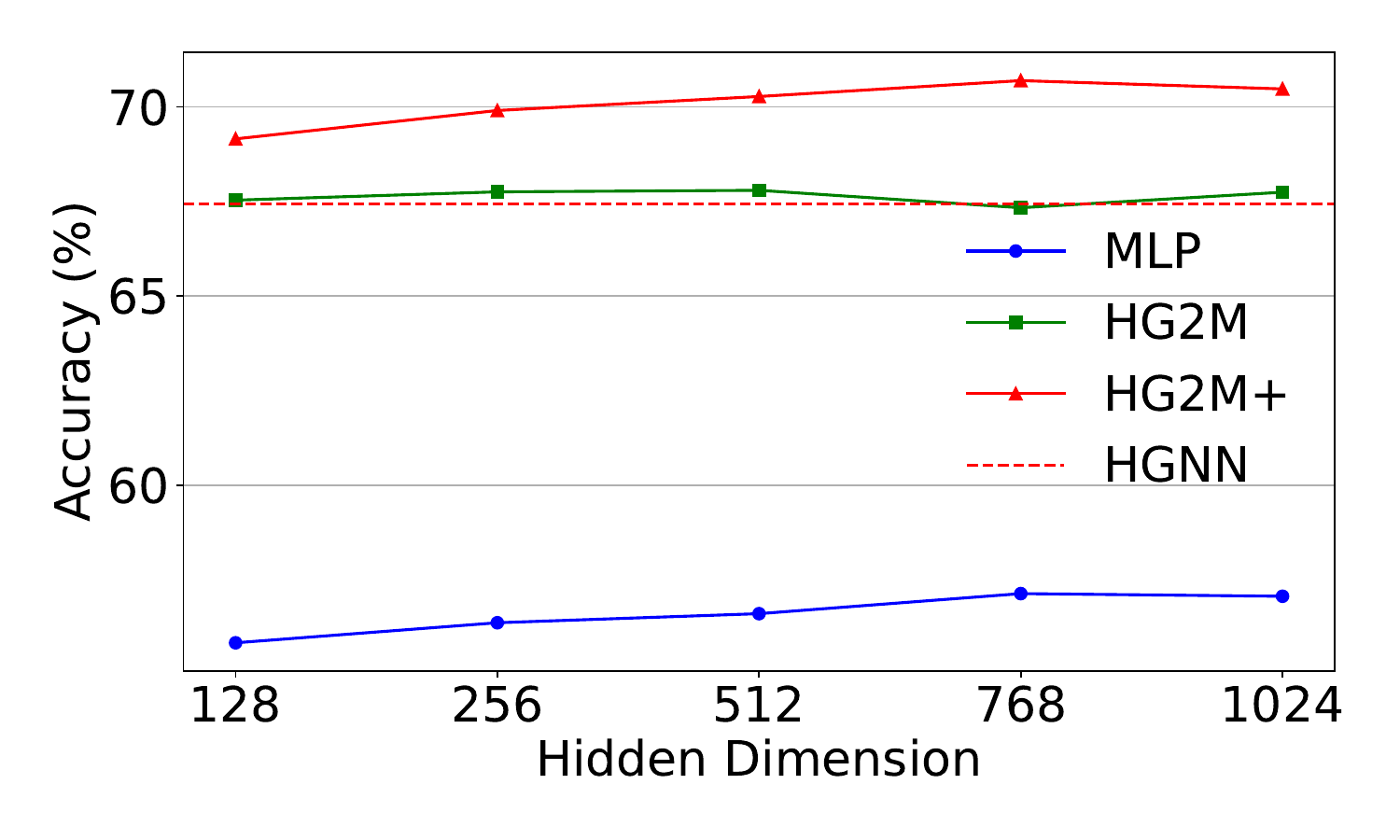}}
    \caption{Impact of model width on the \textit{prod} performance. The teacher HGNN is RSAGE with 128-dim hidden layers.}\label{Fig: Param Hidden Dim}
\end{figure*}

\subsection{How do HG2Ms perform with noisy node features? (RQ7)}\label{Sec: Noisy Node Features}
Considering that HG2Ms utilize only node features as input, they may be susceptible to feature noise and could underperform when the labels are unrelated to the node features. Therefore, we evaluate the robustness of HG2Ms with regards to different noise levels across TMDB, CroVal, and IGB-549K-19 datasets. Specifically, we introduce different levels of Gaussian noises to node features by replacing $\boldsymbol{X}_t$ with $(1-\alpha)\boldsymbol{X}_t+\alpha \epsilon$, where $\epsilon$ is independent isotropic Gaussian noise, and $\alpha \in [0, 1]$ controls the noise level. As depicted in Figure \ref{Fig: Param Noise Ratio}, HG2Ms achieve comparable or improved performance compared to teacher HGNNs across different noise levels. Moreover, HG2Ms not only outperform MLPs but also exhibit slower performance degradation as $\alpha$ increases. When $\alpha$ approaches 1, the input features and node labels will become independent corresponding to the extreme case discussed in Section \ref{Sec: Theoretical Analysis}. In this scenario, HG2M+ continues to perform as well as HGNNs, while MLPs perform poorly.


\subsection{How do different hyperparameters affect HG2Ms? (RQ8)}

\subsubsection{\textbf{Inductive Split Rate}}\label{Sec: Inductive Split Rate}
In Table \ref{Tab: Production Setting}, we employ a 20-80 split of the test data for inductive evaluation. Here, we conduct an ablation study on the inductive split rate under the production setting across TMDB, CroVal, and IGB-549K-19 datasets. 
Figure \ref{Fig: Param Split Rate} shows that altering the inductive:transductive ratio in the production setting does not affect the accuracy much. The performance trend of HG2Ms aligns with that of teacher HGNNs as the split rate changes.
We only consider rates up to 50-50 since having 50\% or more inductive nodes is exceedingly rare in practical scenarios. In cases where a substantial influx of new data occurs, practitioners can choose to retrain the model on the entire dataset before deployment.


\subsubsection{\textbf{Reliable Node Proportion}}\label{Sec: Reliable Node Proportion}
The proportion $p$ of selected unlabelled reliable nodes plays a critical in RND and RMPD. We conduct experiments across TMDB, CroVal, and IGB-549K-19 datasets and report transductive performance to present how the performance of HG2M+ and the accuracy of selected unlabelled reliable nodes vary with $p$. Intuitively, as $p$ decreases, RND tends to select nodes with higher prediction confidence and lower prediction uncertainty, thereby increasing the accuracy of selected unlabelled reliable nodes, as depicted in Figure \ref{Fig: Param Proportion P2}. However, Figure \ref{Fig: Param Proportion P1} shows that decreasing $p$ does not consistently improve the performance of HG2M+, as fewer selected nodes imply less supervision during distillation. In our experiments, we set $p$ to 0.9 across all datasets.


\subsubsection{\textbf{Hidden Dimension}}\label{Sec: Hidden Dimension}
Here, we vary the hidden dimension of student MLPs and assess the production performance of HG2Ms to study the impact of model width, as depicted in Figure \ref{Fig: Param Hidden Dim}. On the ArXiv dataset, we observe that as the hidden dimension increases, the performance of HG2Ms initially improves rapidly and then stabilizes, with the performance gap between HG2Ms and HGNNs narrowed. On the TMDB and IGB-549K-19 datasets, HG2M maintains consistent performance, whereas HG2M+ exhibits a slight improvement.


\subsubsection{\textbf{Trade-off Weight $\lambda$}}\label{Sec: Param Lambda}
\begin{figure}[ht]
    \centering
    \subfigure[TMDB]{\includegraphics[width=0.49\linewidth]{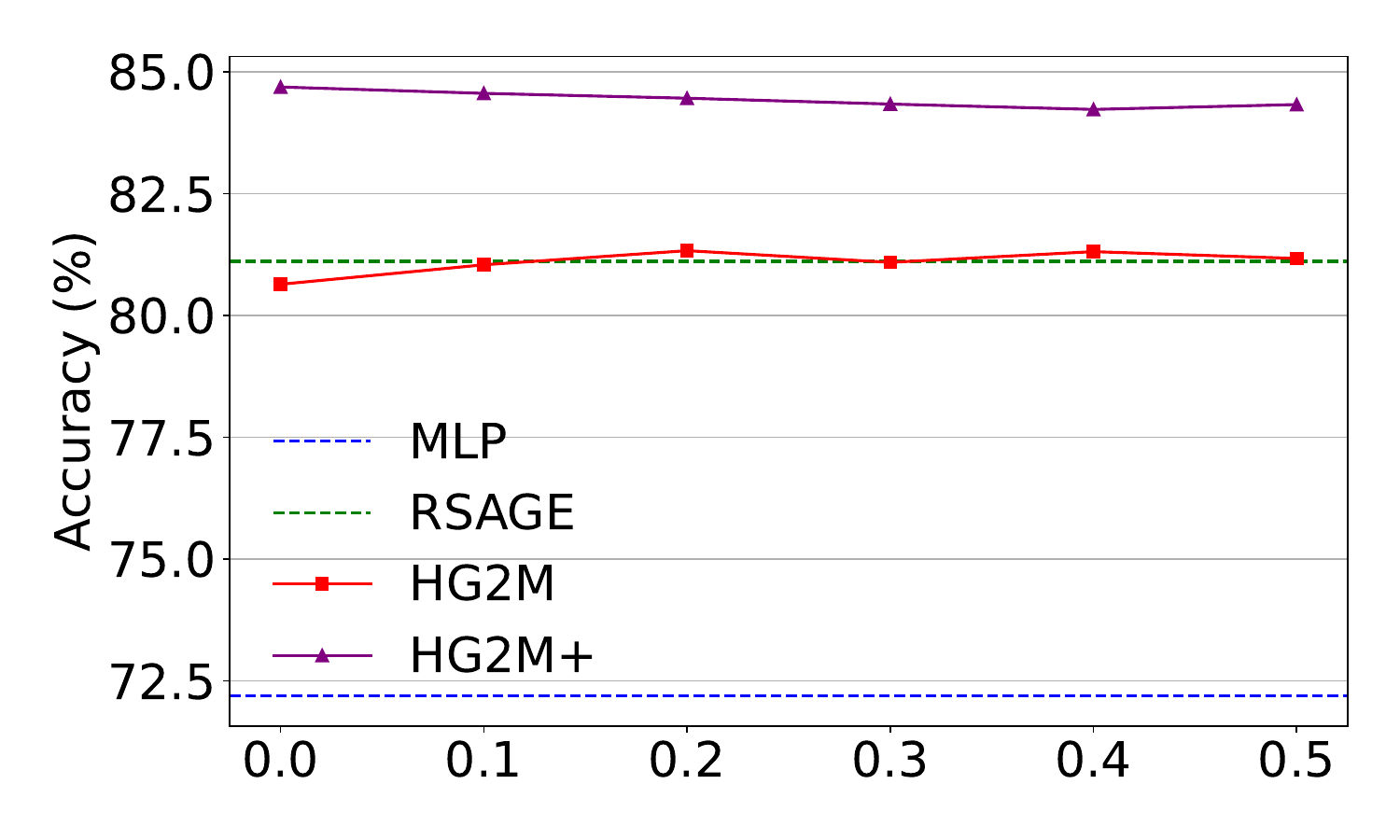}}
    \subfigure[IGB-549K-19]{\includegraphics[width=0.49\linewidth]{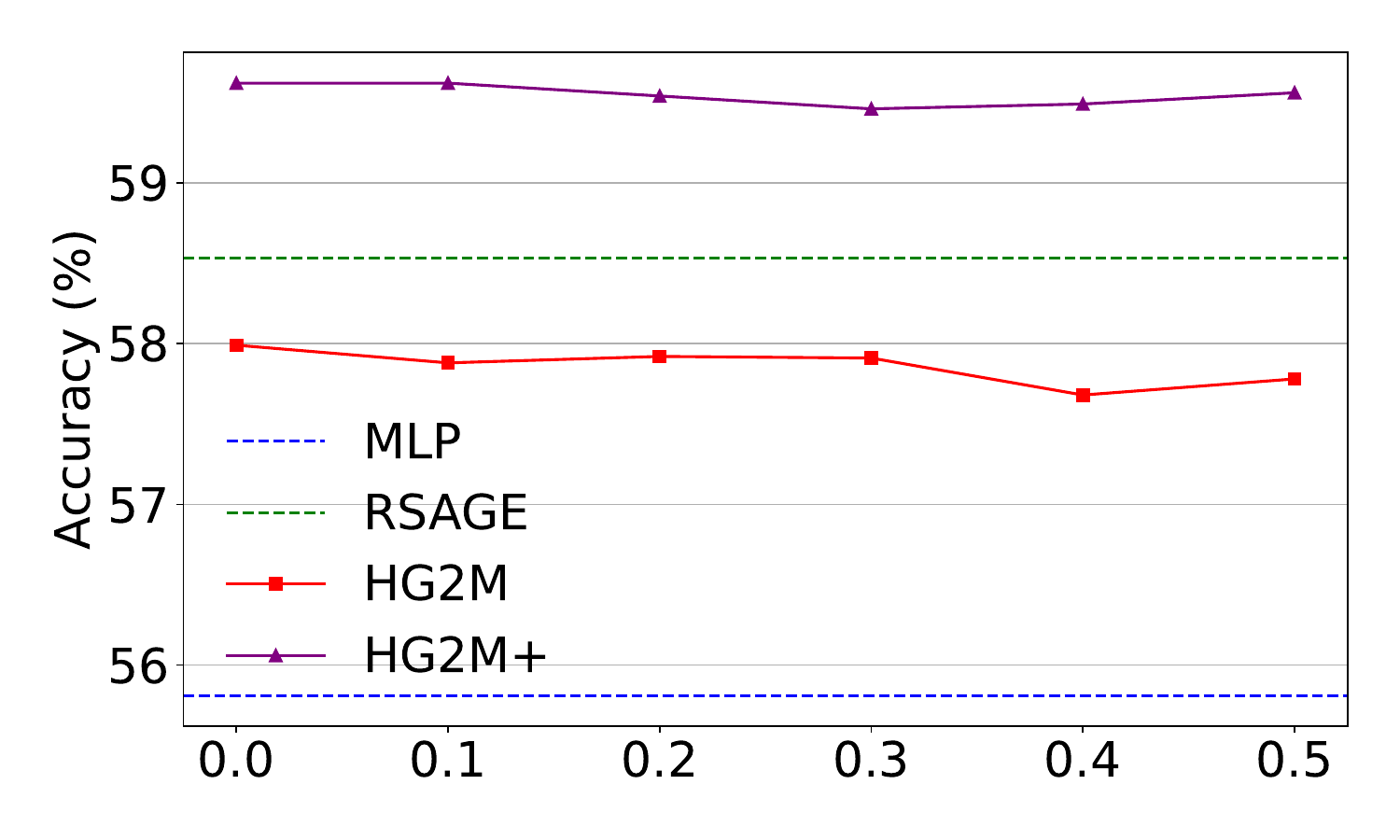}}
    \caption{Transductive Accuracy vs. Trade-off Weight $\lambda$.}\label{Fig: Param Lambda}
\end{figure}

We conduct a sensitivity analysis of $\lambda$ on TMDB and IGB-549K-19. 
As shown in Figure~\ref{Fig: Param Lambda}, increasing $\lambda$ gradually emphasizes the supervised signal, but performance does not significantly improve with larger $\lambda$ values. 
In fact, setting $\lambda=0$, i.e., relying solely on distillation, often achieves comparable or better performance. 
This is consistent with observations in \cite{GLNN, LightHGNN, MGFNN}, which suggest that carefully designed distillation objectives can effectively guide the student model even in the absence of direct supervision. 
Based on this analysis, we set $\lambda=0$ in all our experiments for simplicity and efficiency.

\section{Conclusion}
In this paper, we propose HG2M and HG2M+ to combine both HGNN's superior performance and MLP's efficient inference. HG2M directly trains student MLPs with node features as input and soft labels from teacher HGNNs as targets, and HG2M+ further distills reliable and heterogeneous semantic knowledge into student MLPs through reliable node distillation and reliable meta-path distillation. Experiments conducted on six heterogeneous graph datasets show that HG2Ms can achieve competitive or even better performance than HGNNs and significantly outperform vanilla MLPs. Moreover, HG2Ms demonstrate a 39.81×-379.24× speedup in inference over HGNNs, showing their ability for latency-sensitive deployments.

\noindent\textbf{Limitations and Future Work.}\quad
Our HG2M+ uses feature vectors in RMPD based on attribute, structural, and label similarity, which may not generalize well to all heterogeneous graphs, especially when attributes are sparse or meta-paths lack semantic meaning. Additionally, HG2M and HG2M+ are currently designed for node classification. Extending them to other tasks such as link prediction and graph classification with task-specific distillation strategies is a promising direction for future work.

\section*{Acknowledgment}
This work is partially supported by the National Key Research and Development Program of China (2024YFB2505604), the National Natural Science Foundation of China (62306137), the Australian Research Council under the streams of Future Fellowship (FT210100624), Discovery Project (DP240101108), and Linkage Projects (LP230200892 and LP240200546).

\bibliographystyle{IEEEtran}
\bibliography{main}

\begin{thebibliography}{10}
\providecommand{\url}[1]{#1}
\csname url@samestyle\endcsname
\providecommand{\newblock}{\relax}
\providecommand{\bibinfo}[2]{#2}
\providecommand{\BIBentrySTDinterwordspacing}{\spaceskip=0pt\relax}
\providecommand{\BIBentryALTinterwordstretchfactor}{4}
\providecommand{\BIBentryALTinterwordspacing}{\spaceskip=\fontdimen2\font plus
\BIBentryALTinterwordstretchfactor\fontdimen3\font minus \fontdimen4\font\relax}
\providecommand{\BIBforeignlanguage}[2]{{%
\expandafter\ifx\csname l@#1\endcsname\relax
\typeout{** WARNING: IEEEtran.bst: No hyphenation pattern has been}%
\typeout{** loaded for the language `#1'. Using the pattern for}%
\typeout{** the default language instead.}%
\else
\language=\csname l@#1\endcsname
\fi
#2}}
\providecommand{\BIBdecl}{\relax}
\BIBdecl

\bibitem{HGSurvey}
C.~Yang, Y.~Xiao, Y.~Zhang, Y.~Sun, and J.~Han, ``Heterogeneous network representation learning: A unified framework with survey and benchmark,'' \emph{IEEE Transactions on Knowledge and Data Engineering}, vol.~34, no.~10, pp. 4854--4873, 2020.

\bibitem{MV-URL}
W.~Wang, H.~Yin, X.~Du, W.~Hua, Y.~Li, and Q.~V.~H. Nguyen, ``Online user representation learning across heterogeneous social networks,'' in \emph{International ACM SIGIR Conference on Research and Development in Information Retrieval}, 2019, pp. 545--554.

\bibitem{OAG}
F.~Zhang, X.~Liu, J.~Tang, Y.~Dong, P.~Yao, J.~Zhang, X.~Gu, Y.~Wang, E.~Kharlamov, B.~Shao \emph{et~al.}, ``Oag: Linking entities across large-scale heterogeneous knowledge graphs,'' \emph{IEEE Transactions on Knowledge and Data Engineering}, vol.~35, no.~9, pp. 9225--9239, 2022.

\bibitem{HERec}
C.~Shi, B.~Hu, W.~X. Zhao, and S.~Y. Philip, ``Heterogeneous information network embedding for recommendation,'' \emph{IEEE Transactions on Knowledge and Data Engineering}, vol.~31, no.~2, pp. 357--370, 2018.

\bibitem{MRGAT}
Y.~Zhao, H.~Zhou, A.~Zhang, R.~Xie, Q.~Li, and F.~Zhuang, ``Connecting embeddings based on multiplex relational graph attention networks for knowledge graph entity typing,'' \emph{IEEE Transactions on Knowledge and Data Engineering}, vol.~35, no.~5, pp. 4608--4620, 2022.

\bibitem{DeepMAPS}
A.~Ma, X.~Wang, J.~Li, C.~Wang, T.~Xiao, Y.~Liu, H.~Cheng, J.~Wang, Y.~Li, Y.~Chang \emph{et~al.}, ``Single-cell biological network inference using a heterogeneous graph transformer,'' \emph{Nature Communications}, vol.~14, no.~1, p. 964, 2023.

\bibitem{RGCN}
M.~Schlichtkrull, T.~N. Kipf, P.~Bloem, R.~van den Berg, I.~Titov, and M.~Welling, ``Modeling relational data with graph convolutional networks,'' in \emph{The Semantic Web}.\hskip 1em plus 0.5em minus 0.4em\relax Springer, 2018, pp. 593--607.

\bibitem{ieHGCN}
Y.~Yang, Z.~Guan, J.~Li, W.~Zhao, J.~Cui, and Q.~Wang, ``Interpretable and efficient heterogeneous graph convolutional network,'' \emph{IEEE Transactions on Knowledge and Data Engineering}, vol.~35, no.~2, pp. 1637--1650, 2021.

\bibitem{SimpleHGN}
Q.~Lv, M.~Ding, Q.~Liu, Y.~Chen, W.~Feng, S.~He, C.~Zhou, J.~Jiang, Y.~Dong, and J.~Tang, ``Are we really making much progress? revisiting, benchmarking and refining heterogeneous graph neural networks,'' in \emph{ACM SIGKDD Conference on Knowledge Discovery \& Data Mining}, 2021, pp. 1150--1160.

\bibitem{HGAMLP}
Y.~Liang, W.~Zhang, Z.~Sheng, L.~Yang, J.~Jiang, Y.~Tong, and B.~Cui, ``Hgamlp: Heterogeneous graph attention mlp with de-redundancy mechanism,'' in \emph{International Conference on Data Engineering}, 2024, pp. 2779--2791.

\bibitem{HAN}
X.~Wang, H.~Ji, C.~Shi, B.~Wang, Y.~Ye, P.~Cui, and P.~S. Yu, ``Heterogeneous graph attention network,'' in \emph{The Web Conference}, 2019, pp. 2022--2032.

\bibitem{MAGNN}
X.~Fu, J.~Zhang, Z.~Meng, and I.~King, ``Magnn: Metapath aggregated graph neural network for heterogeneous graph embedding,'' in \emph{The Web Conference}, 2020, pp. 2331--2341.

\bibitem{HPN}
H.~Ji, X.~Wang, C.~Shi, B.~Wang, and S.~Y. Philip, ``Heterogeneous graph propagation network,'' \emph{IEEE Transactions on Knowledge and Data Engineering}, vol.~35, no.~1, pp. 521--532, 2021.

\bibitem{NS4GC}
Y.~Liu, X.~Gao, T.~He, T.~Zheng, J.~Zhao, and H.~Yin, ``Reliable node similarity matrix guided contrastive graph clustering,'' \emph{IEEE Transactions on Knowledge and Data Engineering}, vol.~36, no.~12, pp. 9123--9135, 2024.

\bibitem{Paths2Pair}
J.~Hang, Z.~Hong, X.~Feng, G.~Wang, G.~Yang, F.~Li, X.~Song, and D.~Zhang, ``Paths2pair: Meta-path based link prediction in billion-scale commercial heterogeneous graphs,'' in \emph{ACM SIGKDD Conference on Knowledge Discovery \& Data Mining}, 2024, pp. 5082--5092.

\bibitem{IGB}
A.~Khatua, V.~S. Mailthody, B.~Taleka, T.~Ma, X.~Song, and W.-m. Hwu, ``Igb: Addressing the gaps in labeling, features, heterogeneity, and size of public graph datasets for deep learning research,'' in \emph{ACM SIGKDD Conference on Knowledge Discovery \& Data Mining}, 2023, pp. 4284--4295.

\bibitem{GLNN}
S.~Zhang, Y.~Liu, Y.~Sun, and N.~Shah, ``Graph-less neural networks: Teaching old {MLP}s new tricks via distillation,'' in \emph{International Conference on Learning Representations}, 2022.

\bibitem{NOSMOG}
Y.~Tian, C.~Zhang, Z.~Guo, X.~Zhang, and N.~Chawla, ``Learning {MLP}s on graphs: A unified view of effectiveness, robustness, and efficiency,'' in \emph{International Conference on Learning Representations}, 2023.

\bibitem{KRD}
L.~Wu, H.~Lin, Y.~Huang, and S.~Z. Li, ``Quantifying the knowledge in gnns for reliable distillation into mlps,'' in \emph{International Conference on Machine Learning}.\hskip 1em plus 0.5em minus 0.4em\relax PMLR, 2023, pp. 37\,571--37\,581.

\bibitem{VQGraph}
L.~Yang, Y.~Tian, M.~Xu, Z.~Liu, S.~Hong, W.~Qu, W.~Zhang, B.~CUI, M.~Zhang, and J.~Leskovec, ``{VQG}raph: Rethinking graph representation space for bridging {GNN}s and {MLP}s,'' in \emph{International Conference on Learning Representations}, 2024.

\bibitem{KD}
G.~Hinton, O.~Vinyals, and J.~Dean, ``Distilling the knowledge in a neural network,'' \emph{arXiv preprint arXiv:1503.02531}, 2015.

\bibitem{GCN}
T.~N. Kipf and M.~Welling, ``Semi-supervised classification with graph convolutional networks,'' in \emph{International Conference on Learning Representations}, 2017.

\bibitem{GAT}
P.~Veličković, G.~Cucurull, A.~Casanova, A.~Romero, P.~Liò, and Y.~Bengio, ``Graph attention networks,'' in \emph{International Conference on Learning Representations}, 2018.

\bibitem{RKD-MLP}
Q.~Tan, D.~Zha, N.~Liu, S.-H. Choi, L.~Li, R.~Chen, and X.~Hu, ``Double wins: Boosting accuracy and efficiency of graph neural networks by reliable knowledge distillation,'' in \emph{ICDM}, 2023, pp. 1343--1348.

\bibitem{FF-G2M}
L.~Wu, H.~Lin, Y.~Huang, T.~Fan, and S.~Z. Li, ``Extracting low-/high-frequency knowledge from graph neural networks and injecting it into mlps: An effective gnn-to-mlp distillation framework,'' in \emph{Proceedings of the AAAI Conference on Artificial Intelligence}, vol.~37, no.~9, 2023, pp. 10\,351--10\,360.

\bibitem{LLP}
Z.~Guo, W.~Shiao, S.~Zhang, Y.~Liu, N.~V. Chawla, N.~Shah, and T.~Zhao, ``Linkless link prediction via relational distillation,'' in \emph{International Conference on Machine Learning}.\hskip 1em plus 0.5em minus 0.4em\relax PMLR, 2023, pp. 12\,012--12\,033.

\bibitem{MuGSI}
T.~Yao, J.~Sun, D.~Cao, K.~Zhang, and G.~Chen, ``Mugsi: Distilling gnns with multi-granularity structural information for graph classification,'' in \emph{The Web Conference}, 2024, pp. 709--720.

\bibitem{LightHGNN}
Y.~Feng, Y.~Luo, S.~Ying, and Y.~Gao, ``Light{HGNN}: Distilling hypergraph neural networks into {MLP}s for 100x faster inference,'' in \emph{International Conference on Learning Representations}, 2024.

\bibitem{UPS}
M.~N. Rizve, K.~Duarte, Y.~S. Rawat, and M.~Shah, ``In defense of pseudo-labeling: An uncertainty-aware pseudo-label selection framework for semi-supervised learning,'' in \emph{International Conference on Learning Representations}, 2021.

\bibitem{RDD}
W.~Zhang, X.~Miao, Y.~Shao, J.~Jiang, L.~Chen, O.~Ruas, and B.~Cui, ``Reliable data distillation on graph convolutional network,'' in \emph{Proceedings of the 2020 ACM SIGMOD international conference on management of data}, 2020, pp. 1399--1414.

\bibitem{CPL}
B.~Wang, J.~Li, Y.~Liu, J.~Cheng, Y.~Rong, W.~Wang, and F.~Tsung, ``Deep insights into noisy pseudo labeling on graph data,'' in \emph{Neural Information Processing Systems}, 2023.

\bibitem{HeCo}
X.~Wang, N.~Liu, H.~Han, and C.~Shi, ``Self-supervised heterogeneous graph neural network with co-contrastive learning,'' in \emph{ACM SIGKDD Conference on Knowledge Discovery \& Data Mining}, 2021, pp. 1726--1736.

\bibitem{OGB-LSC}
W.~Hu, M.~Fey, H.~Ren, M.~Nakata, Y.~Dong, and J.~Leskovec, ``{OGB}-{LSC}: A large-scale challenge for machine learning on graphs,'' in \emph{Neural Information Processing Systems}, 2021.

\bibitem{GAMLP}
L.~Chen, Z.~Chen, and J.~Bruna, ``On graph neural networks versus graph-augmented mlps,'' in \emph{International Conference on Learning Representations}, 2021.

\bibitem{qin2019rethinking}
Z.~Qin, D.~Kim, and T.~Gedeon, ``Rethinking softmax with cross-entropy: Neural network classifier as mutual information estimator,'' \emph{arXiv preprint arXiv:1911.10688}, 2019.

\bibitem{AN2VEC}
S.~Lerique, J.~L. Abitbol, and M.~Karsai, ``Joint embedding of structure and features via graph convolutional networks,'' \emph{Applied Network Science}, vol.~5, pp. 1--24, 2020.

\bibitem{HTAG}
Y.~Liu, Q.~Xie, J.~Shi, J.~Shen, and T.~He, ``Multi-scale heterogeneous text-attributed graph datasets from diverse domains,'' in \emph{Companion Proceedings of the ACM Web Conference 2025}, 2025.

\bibitem{MiniLM}
W.~Wang, F.~Wei, L.~Dong, H.~Bao, N.~Yang, and M.~Zhou, ``Minilm: Deep self-attention distillation for task-agnostic compression of pre-trained transformers,'' \emph{Neural Information Processing Systems}, vol.~33, pp. 5776--5788, 2020.

\bibitem{Sentence-BERT}
N.~Reimers and I.~Gurevych, ``Sentence-{BERT}: Sentence embeddings using {S}iamese {BERT}-networks,'' in \emph{EMNLP-IJCNLP}.\hskip 1em plus 0.5em minus 0.4em\relax Association for Computational Linguistics, 2019, pp. 3982--3992.

\bibitem{RoBERTa}
Y.~Liu, M.~Ott, N.~Goyal, J.~Du, M.~Joshi, D.~Chen, O.~Levy, M.~Lewis, L.~Zettlemoyer, and V.~Stoyanov, ``Roberta: A robustly optimized bert pretraining approach,'' \emph{arXiv preprint arXiv:1907.11692}, 2019.

\bibitem{GraphSAGE}
W.~Hamilton, Z.~Ying, and J.~Leskovec, ``Inductive representation learning on large graphs,'' \emph{Neural Information Processing Systems}, vol.~30, 2017.

\bibitem{RGAT}
D.~Busbridge, D.~Sherburn, P.~Cavallo, and N.~Y. Hammerla, ``Relational graph attention networks,'' \emph{arXiv preprint arXiv:1904.05811}, 2019.

\bibitem{Adam}
D.~P. Kingma and J.~Ba, ``Adam: A method for stochastic optimization,'' in \emph{International Conference on Learning Representations}, 2015.

\bibitem{MGFNN}
Y.~Liu, Z.~Tao, X.~Zhao, J.~Zhao, T.~Zheng, and T.~He, ``Learning accurate, efficient, and interpretable mlps on multiplex graphs via node-wise multi-view ensemble distillation,'' in \emph{International Conference on Database Systems for Advanced Applications}.\hskip 1em plus 0.5em minus 0.4em\relax Springer, 2025.

\bibitem{HIRE}
J.~Liu, T.~Zheng, and Q.~Hao, ``Hire: Distilling high-order relational knowledge from heterogeneous graph neural networks,'' \emph{Neurocomputing}, vol. 507, pp. 67--83, 2022.

\bibitem{TGS}
L.~Wu, H.~Lin, Z.~Gao, G.~Zhao, and S.~Z. Li, ``A teacher-free graph knowledge distillation framework with dual self-distillation,'' \emph{IEEE Transactions on Knowledge and Data Engineering}, 2024.

\bibitem{TeKAP}
M.~I. Hossain, S.~Akhter, C.~S. Hong, and E.-N. Huh, ``Single teacher, multiple perspectives: Teacher knowledge augmentation for enhanced knowledge distillation,'' in \emph{The Thirteenth International Conference on Learning Representations}, 2025.

\end{thebibliography}

\end{document}